\documentclass[sigconf]{acmart}

\AtBeginDocument{%
  }

\copyrightyear{2025}
\acmYear{2025}
\setcopyright{cc}
\setcctype{by}
\acmConference[MICRO '25]{58th IEEE/ACM International Symposium on Microarchitecture}{October 18--22, 2025}{Seoul, Republic of Korea}
\acmBooktitle{58th IEEE/ACM International Symposium on Microarchitecture (MICRO '25), October 18--22, 2025, Seoul, Republic of Korea}
\acmDOI{10.1145/3725843.3756118}
\acmISBN{979-8-4007-1573-0/25/10}

\usepackage{xspace}
\usepackage{subcaption}
\usepackage{graphics} 
\usepackage{graphicx,pifont}
\usepackage{mathtools}
\usepackage{multirow}
\usepackage{textcomp}
\usepackage{enumitem}
\usepackage[font=small]{caption}
\usepackage[ruled,noline,noend,linesnumbered]{algorithm2e}

\definecolor{mygray}{rgb}{0.45, 0.45, 0.45}
\definecolor{highcolor}{RGB}{34,139,34} 
\definecolor{lowcolor}{RGB}{220,20,60}

\newcommand{\putsec}[2]{\vspace{-0.00in} \section{#2}\label{sec:#1}}
\newcommand{\putssec}[2]{\vspace{-0.00in}\subsection{#2}\label{ssec:#1}}

\newcommand{\secref}[1]{Section~\ref{sec:#1}}
\newcommand{\ssecref}[1]{Section~\ref{ssec:#1}}

\newcommand{\myparagraph}[1]{\vspace{0.02in}\noindent\textbf{#1}}

\newcommand{\figref}[1]{Figure~\ref{fig:#1}}
\newcommand{\tabref}[1]{Table~\ref{tab:#1}}
\newcommand{\eqnref}[1]{Equation~\ref{eqn:#1}}

\newcommand{\multiline}[2]{\begin{tabular}{@{}c@{}}#1 \\ #2\end{tabular}}
\newcommand{\multilineleft}[2]{\begin{tabular}{@{}l@{}}#1 \\ #2\end{tabular}}

\SetAlgoCaptionSeparator{}
\SetInd{0.78em}{0.01em}
\newcommand*\mycomment[1]{\textcolor{mygray}{\small$\triangleright$\itshape\,#1\;}}

\newcommand*\mycirc[1]{%
  \kern-0.88pt
  \raisebox{-1.23pt}{\scalebox{1.2}{\ding{\numexpr201+#1\relax}}}%
  \kern-0.88pt
}

\newcommand{\low}{\LARGE\textcolor{lowcolor}{\ding{55}}}
\newcommand{\high}{\LARGE\textcolor{highcolor}{\ding{51}}}

\newcommand*\bstrut[1]{\rule[-#1]{0pt}{#1}}

\newcommand{\name}{{MX+\xspace}}
\newcommand{\amxfp}{{A-MXFP4+\xspace}}
\newcommand{\mm}[2]{\ensuremath{{#1}_\mathrm{#2}}}
\newcommand{\emax}{\mm{e}{max}}
\newcommand{\bmh}{\mm{BM}{H}}
\newcommand{\bml}{\mm{BM}{L}}

\begin{document}

\title{MX+: Pushing the Limits of Microscaling Formats for Efficient Large Language Model Serving}
\author{Jungi Lee}
\orcid{0009-0004-5629-6258}
\affiliation{
  \institution{Seoul National University}
  \city{Seoul}
  \country{Republic of Korea}
}
\email{jungi.lee@snu.ac.kr}

\author{Junyong Park}
\orcid{0009-0003-4228-1138}
\affiliation{
  \institution{Seoul National University}
  \city{Seoul}
  \country{Republic of Korea}
}
\email{junyong.park@snu.ac.kr}

\author{Soohyun Cha}
\orcid{0009-0007-8624-9913}
\affiliation{
  \institution{Seoul National University}
  \city{Seoul}
  \country{Republic of Korea}
}
\email{soohyun.cha@snu.ac.kr}

\author{Jaehoon Cho}
\orcid{0009-0003-6853-368X}
\affiliation{
  \institution{Seoul National University}
  \city{Seoul}
  \country{Republic of Korea}
}
\email{jaehoon.cho@snu.ac.kr}

\author{Jaewoong Sim}
\orcid{0000-0002-0403-9928}
\affiliation{
  \institution{Seoul National University}
  \city{Seoul}
  \country{Republic of Korea}
}
\email{jaewoong@snu.ac.kr}

\renewcommand{\shortauthors}{Jungi Lee, Junyong Park, Soohyun Cha, Jaehoon Cho, Jaewoong Sim}




\begin{abstract} 
  Reduced-precision data formats are crucial for cost-effective serving of
  large language models (LLMs). While numerous reduced-precision formats have
  been introduced thus far, they often require intrusive modifications to the
  software frameworks or are rather unconventional for widespread adoption
  across hardware vendors.
  In this paper, we instead focus on recent \emph{industry-driven} variants of
  block floating-point (BFP) formats and conduct a comprehensive analysis to
  push their limits for efficient LLM serving. 
  Our analysis shows that existing ultra low-bit BFP variants struggle to
  provide reasonable language model performance due to outlier values in
  blocks.
  To address the outliers with BFPs, we propose \name{}, a cost-effective and
  non-intrusive extension designed for seamless integration into the
  microscaling (MX) formats.
  \name{} builds on the key insight that the outlier does not need to use its
  exponent field in the element data type, which allows us to \emph{repurpose}
  the exponent field as an extended mantissa to increase the precision of the
  outlier element.
  Our evaluation shows that \name{} achieves significantly higher model
  performance compared to the 4-bit MX format (MXFP4) with negligible storage
  overhead and slowdown, thus offering a compelling alternative to MXFP4 or
  MXFP6 for efficient LLM inference.

\end{abstract}



\maketitle

\putsec{intro}{Introduction}

Emerging services that leverage large language models (LLMs)---such as
human-like chatbots and programming assistants---are increasingly impacting our
daily lives, making LLMs crucial workloads for both software and hardware
vendors today.
Efficiently serving LLMs, however, presents significant challenges due to their
substantial demand for compute and memory resources.
In recent years, numerous low-bit quantization schemes and reduced-precision
data formats have been introduced to alleviate compute and memory overheads.
Yet, they typically require non-negligible changes to the software codebase for
additional operations---such as per-channel scaling~\cite{xia:lin22} and
channel grouping~\cite{tender}---or diverge too far from conventional integer
or floating-point representations to be broadly adopted across computing
platforms~\cite{guo:tan23,guo:che22}.

To address these challenges, industry-leading companies, including AMD, Arm,
Intel, Meta, Microsoft, NVIDIA, and Qualcomm, have recently collaborated to
standardize an open and interoperable family of data formats and introduced
Microscaling (MX) data formats~\cite{mx-ocp}, which build upon block
floating-point (BFP) representations.
However, there has been limited study on their effectiveness in serving LLMs or
on comparisons with other BFP variants from the industry, such as Microsoft
Floating Point (MSFP)~\cite{dar:lo20} or shared microexponents~\cite{smx}.

In this paper, we begin by investigating \emph{industry-driven} BFP variants
and understanding their implications for low-bit LLM inference. Our analysis
shows that MX achieves better language model performance than other BFP
variants that employ similar bit widths. However, employing the 4-bit MX format
(i.e., MXFP4) for \emph{both} weights and activations results in a substantial
reduction in model performance, limiting the maximum benefits of the MX
formats. We observe that this is due to a small number of values whose
magnitudes are significantly larger than others in LLM activations (called
\emph{outliers}), which undergo large quantization errors when converted into
MXFP4.

To effectively address the outliers in BFPs, we propose the \name{} format, a
cost-effective and non-intrusive extension to MX, which enables LLM serving
with \emph{both} 4-bit weights and activations.
The \name{} design builds on two key insights.
First, in MX formats, the exponent of the largest magnitude value in a
block is used for determining a \emph{shared} scale, allowing the natural
identification of an outlier element and its position within the block
\emph{without} additional computation or extra hardware logic.
Second, for the outlier element in an MX block, we do not need to store its
\emph{own} exponent as it is \emph{always} set to the maximum representable
exponent of the element data type. This allows us to \emph{repurpose} the
exponent field to store more mantissa bits for the outlier element, which
greatly helps increase the precision of the outlier in low-bit MX formats,
where only one or a few mantissa bits exist (e.g., MXFP4). 

The non-intrusive design of \name{} makes it easy to integrate into LLM
frameworks and computing systems through both software and hardware approaches.
On the software side, we implement \name{} and evaluate its accuracy and
performance across various LLMs using the MX emulation library~\cite{mx-git},
NVIDIA CUTLASS~\cite{cutlass}, and the Triton compiler~\cite{til:kun19}.
Additionally, we propose an approach to reduce the performance overhead of
software-based \name{} integration by incorporating architectural support into
acceleration units such as Tensor Cores in NVIDIA GPUs. 
Our evaluation shows that \name{} offers a significant improvement in model
performance across a range of LLMs, achieving up to a +42.15\% improvement for
the 4-bit MX format, with a negligible slowdown under software integration or
architectural support. 

In summary, this paper makes the following contributions: 
\begin{itemize}
  \item We propose \name{}, a non-intrusive extension to the MX formats.
    \name{} offers improved representation of outliers without additional user
    effort or complexity and achieves high model performance even with low-bit
    quantization.
  \item We integrate \name{} into existing software libraries and demonstrate
    that it incurs a marginal slowdown in LLM inference, \textit{without}
    hardware modifications, particularly when token generation dominates
    inference time.
  \item We present a hardware design that enables direct \name{} computation
    within Tensor Cores without making intrusive modifications to the dot product
    pipeline, delivering near-MX performance while achieving higher model accuracy.
\end{itemize}

\begin{figure}[t]
  \centering
  \includegraphics[width=0.95\columnwidth,clip,trim=0in 0.10in 0.16in 0.00in]{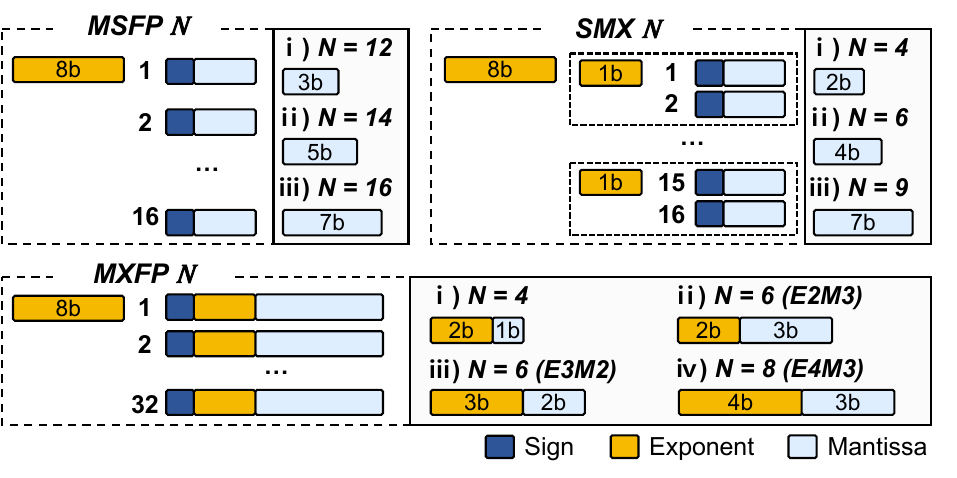}
  \vspace{-0.10in}
  \caption{Industry-driven block floating-point formats.}
  \vspace{-0.16in}
\label{fig:bfp-overview}
\end{figure}

\putsec{back}{Preliminaries: Industry-Driven Block-Based Data Formats}

By mapping weights and activations from higher precision to lower precision
with coarser-grained representations, one can accelerate computation with
simpler yet higher-throughput compute units while also achieving a more
efficient use of memory bandwidth and capacity. For instance, the widely used
uniform, symmetric integer quantization scheme maps a group of $k$
floating-point numbers $x_{f}$ to $b$-bit integers $x_{q}$ using the scale
factor $s$, as shown below: 
\begin{equation*} 
  \begin{aligned} 
    s = \frac{max(|x_{f}|)}{2^{b-1}-1}; \quad x_{q} = round(\frac{x_{f}}{s}).
  \end{aligned} 
\end{equation*}
For integer quantization, dequantization involves multiplying the scale factor
with the integer values, $s{\cdot}x_q$. 

Block floating-point (BFP) formats, such as MSFP~\cite{dar:lo20}, bear some
similarity to conventional integer quantization, but with the scale factor
($s$) \emph{restricted} to powers of two. This restriction enables hardware to
efficiently manage scaling or rescaling at a finer block ($k$) granularity,
thereby allowing for more accurate representations of the original weight and
activation tensors.
\figref{bfp-overview} provides a comparison of several industry-proposed
block-based data formats, which we briefly explain below.

\myparagraph{Microsoft Floating Point.}
Microsoft Floating Point (MSFP) is a variant of BFP formats, which was deployed
in Project Brainwave~\cite{fow:ovt18}. An MSFP block comprises $k$ number of
elements, each with its own sign bit and mantissa bits, and a \emph{shared}
exponent used by all elements in the block. 
In a typical use case of MSFP~\cite{dar:lo20}, for instance, 16 elements in a
floating-point tensor are grouped into a block with an 8-bit \emph{shared}
exponent, which is set to the exponent of the largest absolute value within the
block.
The mantissa of each element is obtained by right-shifting the original
floating-point value by the difference between the shared exponent and its
original exponent; therefore, there are no implicit leading bits in the MSFP
mantissa. Note that MSFP formats are named based on their total bit width; for
instance, MSFP12 has only four bits for the sign and mantissa, resulting in an
average bit width of 4.5 bits per element.

\myparagraph{Shared Microexponents.} 
Shared microexponents (SMX) data formats~\cite{smx} are a recent proposal
similar to MSFP in that scale factors are shared by a group of elements and
restricted to powers of two.\footnote{We denote the shared microexponents data
formats as SMX in this paper to distinguish it from the OCP proposal of
Microscaling (MX) formats.}
However, SMX employs a \emph{multi-level scaling} approach, in contrast to the
single-level scaling in MSFP. In its typical use case of \emph{two-level
scaling}~\cite{smx}, a group of 16 elements ($k_{1}$=16) shares a
\emph{first-level} scale factor $s$, which is an 8-bit shared exponent, while
pairs of elements ($k_{2}$=2) \emph{within} the group form a subgroup that
shares a \emph{second-level} scale (sub-scale) factor $ss$, represented by a
one-bit shared microexponent for each subgroup.

\begin{table}[t]
  \caption{Concrete MX-compliant formats.}
  \vspace{-0.10in}
  \centering
    \resizebox{0.95\columnwidth}{!}{%
      {\fontsize{10}{12}\selectfont
      \renewcommand{\arraystretch}{0.95}
        \begin{tabular}{lcccc}
          \toprule
          \textbf{Name}           & \textbf{Element Data Type}    & \textbf{Bits}        
                                  & \textbf{Block Size ($k$)}     & \textbf{Scale Format}  \\
          \midrule                                                                           
          \multirow{2}{*}{MXFP8}  & E5M2  &\multirow{2}{*}{8}  & 32  & E8M0                \\
                                  & E4M3  &                    & 32  & E8M0                \\ 
          \midrule                                                                                                       
          \multirow{2}{*}{MXFP6}  & E3M2  &\multirow{2}{*}{6}  & 32  & E8M0                \\
                                  & E2M3  &                    & 32  & E8M0                \\ 
          \midrule                                                               
          MXFP4                   & E2M1  &4                   & 32  & E8M0                \\
          \midrule                                                               
          MXINT8                  & INT8  &8                   & 32  & E8M0                \\
          \bottomrule
        \end{tabular}
      }
    }
  \label{tab:mx-formats}
  \vspace{-0.10in}
\end{table}

\myparagraph{Microscaling Formats.} 
Microscaling (MX) formats~\cite{mx-ocp} are another recent proposal developed
in collaboration with multiple industry companies, with the objective of
establishing open and interoperable data formats.
An MX block consists of 32 elements ($k$=32) with a shared scale $X$, which is
an 8-bit shared exponent, similar to the one used in MSFP or SMX. However,
unlike MSFP or SMX, the element data type can be selected from five
floating-point and one integer encodings, as shown in~\tabref{mx-formats}.

The integer data type (i.e., MXINT8) uses a two's complement encoding and has
an implicit scale factor of $2^{-6}$.
For MXFP formats, each private element in an MX block has its \emph{own}
exponent bits, which makes each element effectively a floating-point number.
In MX, the shared exponent and the corresponding scale factor $X$ can be
computed as follows:
\begin{equation} 
  \begin{aligned} 
    \mathrm{shared\_exp} \, =& \, max(\lfloor{log_2(|x_{f}|)}\rfloor) - e_{\mathrm{max}}, \\ 
    {X}                  \, =& \, 2^{\mathrm{shared\_exp}},
  \end{aligned} \label{eqn:mxfp-scale}
\end{equation}
\noindent where \mm{e}{max} is the maximum representable exponent value of the
element data type. For instance, in MXFP4, the element data type is FP4, and
each element has an individual 2-bit exponent with the exponent bias of 1.
Thus, \mm{e}{max} becomes 2 (i.e., $11_2-1$).

\putsec{char}{Serving LLMs with Block-Based Data Formats}

In this section, we first compare the model performance of various LLMs when
using the BFP variants discussed in~\secref{back}. 
We then investigate the root cause of performance degradations with extremely
low-bit formats and discuss how to better exploit low-bit BFP for LLM serving.

\begin{figure}[t]
  \includegraphics[width=\columnwidth,clip,trim=0in 0.00in 0.00in 0.00in]{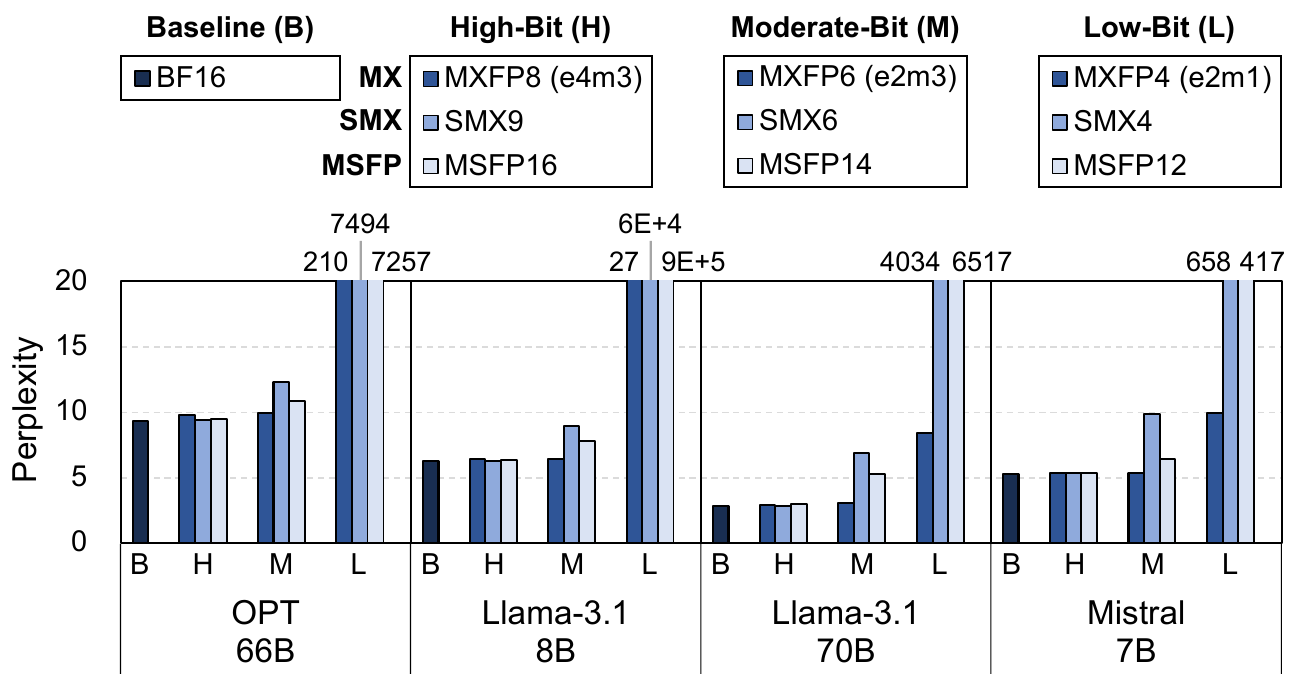}
  \vspace{-0.25in}
  \caption{
           Perplexity with BF16 baseline (B), MSFP, SMX, and MX formats.
           We compare each format with varying bit widths: high (H), moderate
           (M), and low (L).
          }
  \vspace{-0.15in}
  \label{fig:ppl-sweep}
\end{figure}

\putssec{}{Model Performance with BFP Variants}
\figref{ppl-sweep} shows the perplexity of various LLMs across different
industry-driven BFP formats using the WikiText-2 dataset with the sequence
length of 2048.\footnote{Perplexity is a widely used metric to assess the model
performance of generative LLMs; lower values indicate better performance.} 
The baseline (B) uses Bfloat16 (BF16) as the default data format and performs
matrix multiplications and element-wise operations in BF16, except for softmax,
which uses FP32.
For the evaluation with BFP variants, we follow the computation flow outlined
in prior work~\cite{smx,mx-arxiv}; BF16 tensors are converted into MSFP, SMX,
and MX for matrix multiplications, while element-wise operations use the same
precision as the baseline (i.e., BF16 or FP32).
We select the MSFP and SMX formats with average bits per element similar to
those in MXFP4 (L), MXFP6 (M), and MXFP8 (H).
The average bit widths of these BFP variants fall within the ranges of {4
$\leq$ L $\leq$ 4.5}, {6 $\leq$ M $\leq$ 6.5}, and {8.25 $\leq$ H $\leq$ 9}. 

In general, MX outperforms or matches other BFP variants with similar bit
widths.
For the high-bit (H) formats, all BFP variants perform close to the baseline;
while MXFP8 shows slightly higher perplexity than SMX9 or MSFP16, this is due
to its lower average bits per element (8.25, compared to 9 and 8.5 in the other
formats) and the use of reserved NaN representations, which are not supported
by SMX or MSFP.
In the moderate-bit (M) formats, however, SMX6 and MSFP14 begin to diverge,
making them less effective for LLM serving scenarios, while MXFP6 remains close
to the baseline.
This is because, unlike SMX or MSFP, where exponents are shared among some or
all elements in a block, each element in MXFP has its \emph{own} exponent in
addition to the \emph{shared} one, allowing for more fine-grained value
representations.
In addition, MXFP employs an \emph{implicit} leading 1 for normals, with
sub-normals defined similarly to IEEE-754 floating-point formats, resulting in
a larger effective bit width compared to other formats.
However, when using the low-bit MX format (i.e., MXFP4), perplexity also begins
to deviate from the baseline, even with MX. While it still significantly
outperforms SMX4 and MSFP12, this makes the low-bit MX format less suitable for
practical use in LLM serving, despite its potential for substantial bandwidth
savings and computational efficiency.

\begin{figure}[t]
  \includegraphics[width=\columnwidth]{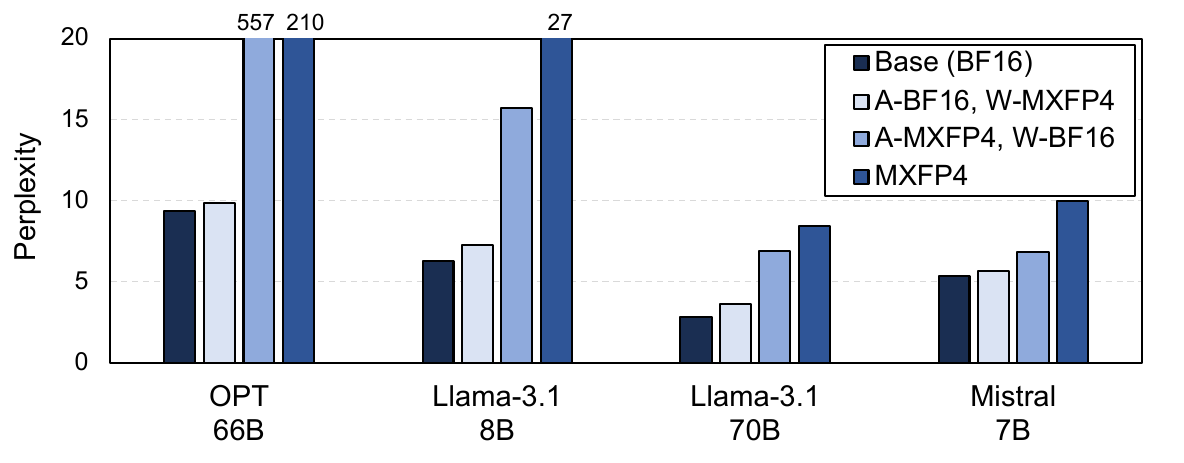}
  \vspace{-0.25in}
  \caption{Perplexity of WikiText-2 across a mix of BF16 and MXFP4.}
  \vspace{-0.15in}
  \label{fig:act-analysis-1}
\end{figure}

\putssec{mxfp4-analysis}{Analysis on Low-Bit MX Format}

To understand the underlying reasons for the reduction in model performance with the
low-bit MX format, we conduct a further analysis on MXFP4.
We first evaluate the perplexity when only either activation tensors (A) or
weight tensors (W) are quantized to MXFP4, while the others use BF16.
\figref{act-analysis-1} shows that quantizing weights (A-BF16, W-MXFP4)
leads to a negligible increase in perplexity, whereas quantizing activations
(A-MXFP4, W-BF16) substantially degrades model performance.
This indicates that although MX employs fine-grained scaling (i.e., 32 elements
per block) to reduce the impact of outlier values in tensors, low-bit MX does
not effectively mitigate this for activation tensors.

\begin{figure}[t]
  \includegraphics[width=\columnwidth]{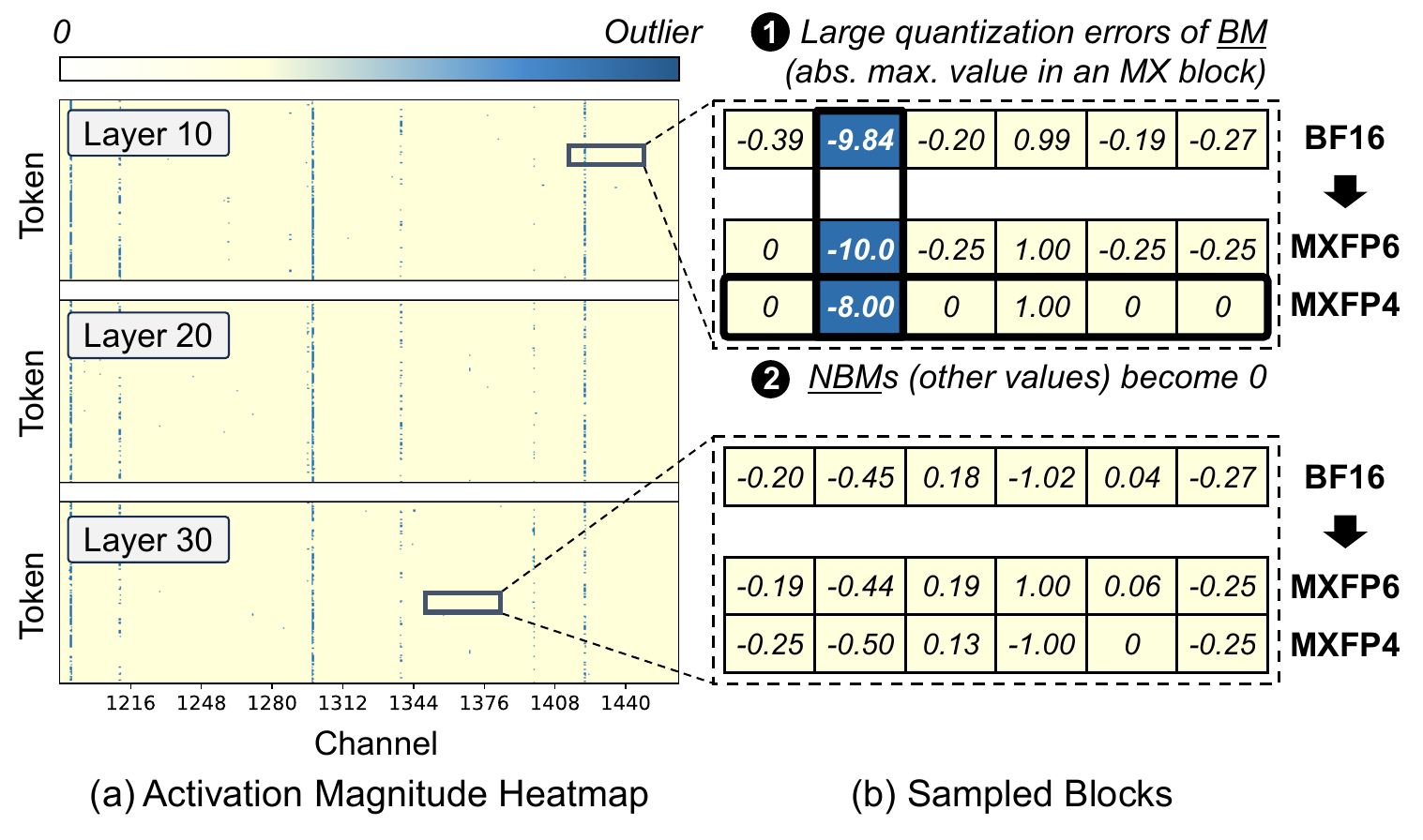}
  \vspace{-0.25in}
  \caption{
           (a) Heatmap of the sample attention input tensors of Llama-3.1-8B.
           (b) BF16 values and their MXFP4 and MXFP6 representations. BF16
           values are rounded to the second decimal place for brevity. 
          }
  \vspace{-0.15in}
  \label{fig:act-analysis-2}
\end{figure}

To gain deeper insight into the root cause, we examine the MX blocks in the
activation tensors.
\figref{act-analysis-2}(a) shows a heatmap of activation magnitudes from
Llama-3.1-8B, while \figref{act-analysis-2}(b) presents two sample blocks with
the original BF16 values and their MXFP4 and MXFP6 representations.
Here, we refer to the \emph{absolute maximum value} in an MX block as
\emph{Block Max (BM)}, while the other values are referred to as
\emph{Non-Block Max (NBM)}.

We observe that blocks in which the BM is significantly larger than the NBMs
(e.g., the upper sampled block)---primarily due to the presence of an outlier
in the MX block---tend to exhibit high quantization errors for two reasons.
First, since MXFP4 allocates only 1 bit for the mantissa, large-magnitude
BMs are susceptible to substantial deviation from their original values upon
quantization~\mbox{(\mycirc{1})}.
Second, the exponent of a BM dictates a block-wide scale factor, as shown
in~\eqnref{mxfp-scale}. As a result, when the BM is large, the shared scale
factor also becomes large, which forces the other smaller elements (i.e., NBMs)
to be represented less precisely due to the shared scale. 
For instance, most NBMs are quantized to zero after being divided by the shared
scale factor~\mbox{(\mycirc{2})}.
In contrast, blocks without outliers (e.g., the lower sampled block) naturally
have relatively low quantization errors.

\begin{figure}[t]
  \includegraphics[width=\columnwidth]{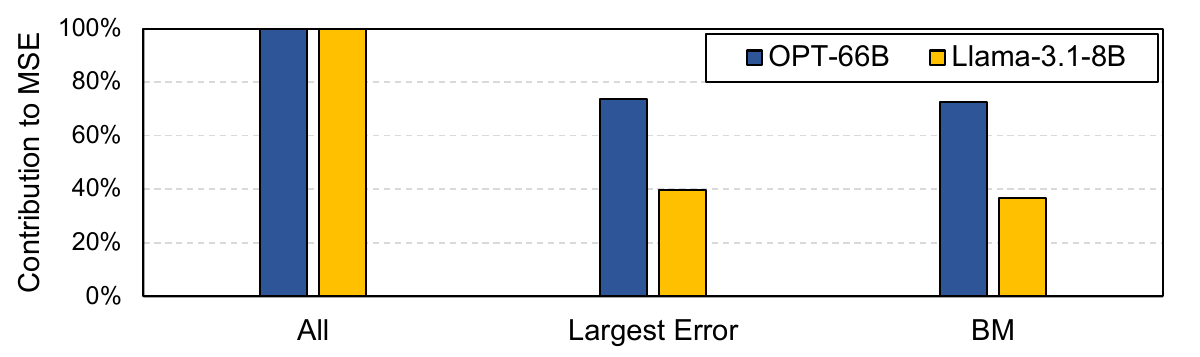}
  \vspace{-0.25in}
  \caption{
           Contribution to MSE (\%) from elements with the largest error in
           each MX block or from BM elements. We use the sampled attention
           input tensor of Layer 16.
          }
  \vspace{-0.20in}
  \label{fig:mse-bm}
\end{figure}

\figref{mse-bm} shows the contribution to MSE (Mean Squared Error) from elements with the largest
quantization error or from BM elements. 
We can see that more accurately representing the BM element in every MX block
can reduce a significant portion of the quantization error. 
While theoretically any of the 32 elements in an MX block could have the
largest error for the block, identifying the largest error element in every
block introduces computational complexity without adding much benefit, as the
BM element is often the largest contributor.

In summary, while it would be ideal to precisely represent all 32 elements in a
block, this is not likely feasible with MXFP4 due to the limited number of
mantissa bits. Instead, our analysis shows that focusing on better
representation of the BM element alone can noticeably help improve model
performance when using the low-bit MX format.

\putsec{mx-plus}{\name{}: Enhancing the MX Formats}

As discussed in~\ssecref{mxfp4-analysis}, the largest magnitude in an MX block
often experiences the highest quantization error among the elements within the
block, which significantly hurts model performance particularly when the MX
block contains an outlier.
In this section, we propose \name{}, a cost-effective extension designed for
seamless integration into the MX formats for low-bit LLM serving.

\putssec{overview}{\name{} Design}

Our \name{} design revolves around three key considerations.
First, it should not interfere with the design goals of MX; the extension needs
to be managed \emph{solely} within conversion kernels or hardware units for
seamless integration with various frameworks, without requiring additional
effort from end-users.
Second, the extension needs to effectively handle outliers in blocks to mitigate
quantization errors while aligning with the MX specification~\cite{mx-ocp}. 
Third, the extension should not introduce large overheads in terms of storage
and runtime latency.

\name{} builds on two key insights. First, the BM element in each MX block
does not need to store its \emph{own} exponent, as it is \emph{always} set to
the maximum representable exponent of the element data type, provided
extremely small magnitude numbers below a threshold are flushed to
zero.
This allows us to safely \emph{repurpose} the exponent field as an
\emph{extended mantissa} to more precisely represent the BM.
Second, the BM element is also \emph{naturally} identified during conversion
from higher precision to MX for computing the shared scale. Thus, no additional
computation is required to identify the BM element in each MX block.

\begin{figure}[t] 
  \centering
  \includegraphics[width=\columnwidth,clip,trim=0.10in 0.00in 0.15in 0in]{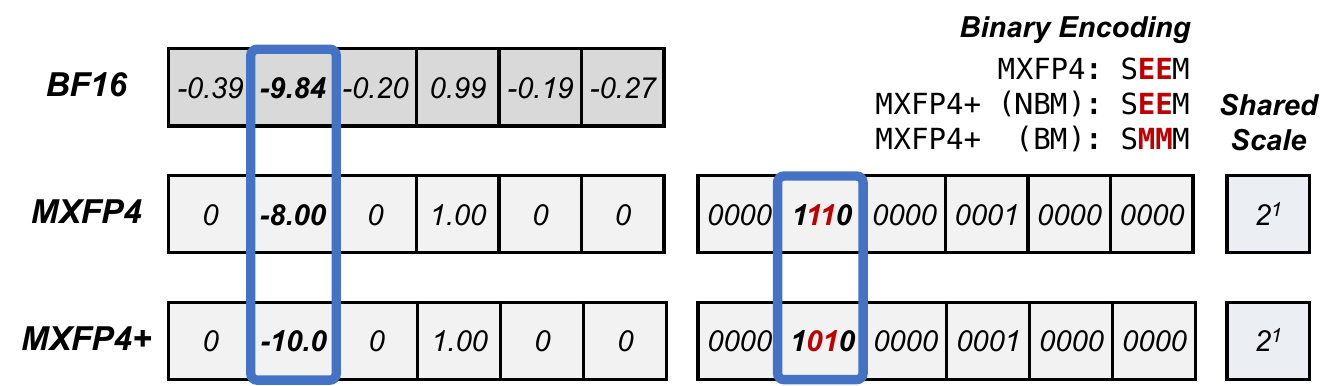}
  \vspace{-0.20in}
  \caption{
           Comparison of MX and MX+ encodings (S: sign, E: exponent, M: mantissa). 
           The BM element is boxed in blue.
           BF16 values are rounded to the second decimal place for brevity.
          }
  \vspace{-0.20in}
  \label{fig:mx-plus-example}
\end{figure}

\figref{mx-plus-example} illustrates an example of binary encoding for MXFP4
and MXFP4+ using the sampled block presented in~\figref{act-analysis-2}. As shown
in the figure, the exponent field of the BM element is always set to the
maximum representable value (i.e., $11_2$) in MXFP4. MXFP4+ repurposes these
bits to store additional mantissa, thereby providing a more accurate
representation of the original BM value. Note that \name{} does not alter the
shared scale.

Similar to prior work~\cite{mx-arxiv}, we flush values with extremely small
magnitudes to zero to simplify conversion and enable the \name{} extension.  
Specifically, if the exponent of the BM ($\lfloor{\mathrm{log_2(BM)}}\rfloor$)
is less than or equal to -127 + \emax{}, we set all elements in the block to
zero.
This is because, in such cases, the shared exponent gets clamped at its lower
bound of -127, which results in the exponent field of the element data type
being set less than \emax{}.
To represent this case, we extend the shared exponent encoding by reserving a
special value: a biased shared exponent of zero indicates that all elements in
the block are zero.

\begin{figure}[t] 
\centering
\includegraphics[width=\columnwidth]{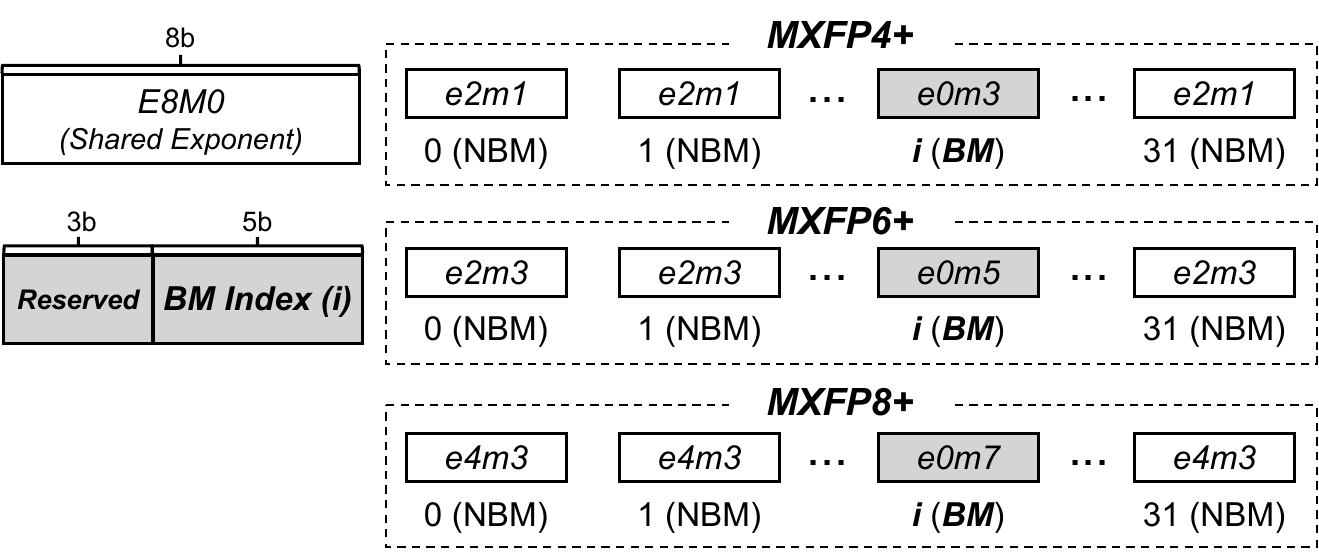}
\vspace{-0.25in}
\caption{
         Data layout of \name{}. We extend the mantissa bits instead of storing
         exponent bits in BM. An additional 8-bit is assigned per block to store
         the index of BM.
        }
\vspace{-0.20in}
\label{fig:data-layout}
\end{figure}

\putssec{layout}{Data Layout of \name{}}

\figref{data-layout} shows the data layout of \name{} for three possible types:
MXFP4+, MXFP6+, and MXFP8+, each of which is an extension to MXFP4, MXFP6
(E2M3), and MXFP8 (E4M3), respectively.
An additional 8 bits are assigned to each MX block, where five bits are used to
store the index of the BM element within the block. 
The remaining three bits are reserved, which can be utilized for further
optimizations or to support future MX specifications that define formats with
block sizes other than 32 elements.

NBM values are converted to conventional MX element data types such as E2M1,
E2M3, and E4M3, while the BM value is stored with more mantissa bits such
as E0M3, E0M5, and E0M7.
We do not explicitly store the exponent of BM since it will always be the
maximum of the given element data type (i.e., 2 for E2M1 and E2M3; 8 for
E4M3), as shown in~\eqnref{mxfp-scale}.
Thus, while using the same bit width as NBMs, BMs are effectively represented
as E2M3, E2M5, and E4M7 for MXFP4+, MXFP6+, and MXFP8+. 

Note that since all elements use the same bit width, MX+ does not lead to
unaligned memory access.
This design also incurs a negligible overhead in terms of compute and memory
costs, as BMs are already identified during conversion to the MX format. The
additional bits increase the average bit width by only 0.25 (e.g., from 4.25 to
4.5 for MXFP4).
Similar to the shared scale~\cite{mx-ocp}, the index metadata does not need
to be stored contiguously with the element data or the shared scale. It can
also be compressed or pruned away for the repeated values.

\putssec{mx++}{Potential Use of Reserved Bits}

While \name{} greatly reduces block-wise quantization error by representing the
BM element more precisely, NBM elements also contribute to quantization error.
As discussed earlier, since the shared scale is determined by the BM, NBMs may
be represented even less precisely than they would be with their own scales.
To show an example of exploiting the reserved bits, we also consider a variant
of \name{}, referred to as MX++, and evaluate its accuracy.

MX++ decouples the shared scale of NBMs from that of the BM by utilizing the
reserved bits, often enabling NBMs to be mapped to a finer quantization grid
compared to \name{}.
Specifically, NBMs employ a shared scale that is smaller than or equal to the
shared scale for the BM, with the difference between their shared
exponents encoded into the reserved three bits in~\figref{data-layout}.
However, directly applying the shared exponent computation
from~\eqnref{mxfp-scale} to NBMs may increase quantization error because NBM
elements could saturate to the maximum magnitude of the element data type after
scaling.
We thus define the smallest feasible shared exponent $e$ for NBMs to avoid
saturation as follows:
\begin{equation} 
  \begin{aligned} 
    \small \nonumber 
    e \, =& \, max_2 \, (\lfloor{log_2(|x_{f}|)}\rfloor) - e_{\mathrm{max}} + 1, \\ 
  \end{aligned} 
  \label{eqn:mx-pp-candidates}
\end{equation}
\noindent where $max_2$ identifies the second-largest exponent in a given MX
block.
Without the offset of 1, the first two terms would represent the shared
exponent of an MX block without the BM (\eqnref{mxfp-scale}), which may
introduce additional error.
Consider the example elements in~\figref{mx-plus-example}. 
Without the offset, $e$ equals -3 ($=-1-2$, where -1 is the exponent of 0.99),
and the value 0.99 is scaled to 7.92 ($=0.99 \div 2{^{-3}}$) and saturated to
the maximum representable value of 6.0 in MXFP4.
With the offset of 1, however, the value is scaled to 3.96 and remains within
the representable range.
The final shared exponent for NBMs is then determined by applying the
clipping function $CLIP(x, \{min, max\})$:
\begin{equation} 
  \begin{aligned} 
    \small \nonumber 
    \mathrm{shared\_exp}_{\mathrm{new}} =
    CLIP(e,\,\{\mathrm{shared\_exp - 7},\,\mathrm{shared\_exp}\}), 
  \end{aligned}
\label{eqn:mx-pp-clipping}
\end{equation} 
\noindent where $\mathrm{shared\_exp}$ and
$\mathrm{shared\_exp_{\mathrm{new}}}$ denote the shared exponents for the BM
and NBMs in MX++.
The lower bound ensures that the difference from the BM's shared exponent
($\mathrm{shared\_exp}$) fits within 3 bits.
The upper bound addresses the case where the exponents of the BM and the largest
NBM are identical and thus $e$ exceeds $\mathrm{shared\_exp}$ due to the offset.
Revisiting the previous example, $\mathrm{shared\_exp_{\mathrm{new}}}$ of -2
enables the NBM value -0.39 to scale to -1.56 and map to -1.5, whereas it was 
previously quantized to zero with the $\mathrm{shared\_exp}$ of 1.

\putsec{sw}{Software Integration of \name{} on GPUs}

MX formats are increasingly integrated into existing DNN acceleration systems
with software and hardware support. In systems lacking compute units for
low-precision element data types in MX, MX blocks are typically converted to a
higher-precision format supported by the
hardware~\cite{wan:ma24,intel-xeon6,til:kun19}.
For example, data stored in MXFP4 can be converted to FP16 for computation on
Intel Granite Rapids via software support~\cite{intel-xeon6}. 
In such scenarios, \name{} can also be easily supported with a minor
modification to the conversion kernel as follows:
\begin{equation} 
\begin{aligned} 
  output_i &= (-1)^{s_i} \times \mathrm{2^{shared\_exp}} \times E_i  \qquad{0 \le i < 32,}\,\, \\ 
  \text{where}~~ 
  E_i &=
  \begin{cases} 
    2^{\emax{}}   \times m_i \quad &\text{if } i = \mathrm{BM~Index,} \quad(\texttt{MX+ Addition})\\
    2^{e_i-b}     \times m_i \quad &\text{otherwise,}
  \end{cases}
\end{aligned} 
\label{eqn:deconversion-kernel}
\end{equation}
where ${i}$ and ${b}$ denote the element index within a block and the exponent
bias of the given element data type. Also, ${s_i}$, ${m_i}$, and ${e_i}$
represent the sign, mantissa, and private exponent of an input element $i$,
respectively.
Note that the mantissa $m_i$ is captured differently between BM and NBM
elements in \name{}. For instance, in MXFP4+, the BM element has three
effective bits for the mantissa, while the NBM elements have only one bit.

When the systems are equipped with compute units that
natively support MX-compliant formats---such as Tensor Cores in the recently
announced NVIDIA Blackwell GPUs~\cite{blackwell}---element data in MX blocks
can be processed directly within the compute unit \emph{without} the need for
format conversion.
In this section, we present an approach to integrate the \name{} extension into
a GPU system that supports MX precision formats \textit{without} requiring any
hardware modifications.
For clarity, we focus on the scenario where activations are represented in
MXFP4+ and weights in MXFP4, a configuration that achieves model performance
comparable to the case where both are in MXFP4+, as we
discuss in~\ssecref{model-perf}.
However, the proposed approach can also be applied when both operands use
MXFP4+ or other formats such as MXFP6+.

\putssec{gpu-challenge}{Challenges of Handling BM on GPUs}

NVIDIA's Tensor Core performs matrix multiply-and-accumulate (MMA) operations
($D = A \times B + C$) for a \emph{predefined} set of matrix tile shapes ($M
\times N \times K$) with specific input and output data types.
Tensor Core MMA operations are exposed to programmers through PTX instructions
such as \textit{wmma.mma}, \textit{mma}, and \textit{wgmma.mma\_async}, which
are translated into device-specific machine code (i.e., SASS instructions) such
as HMMA (half-precision) and IMMA (integer).
We discuss this section based on the following MXFP4 \textit{mma} PTX
instruction without loss of generality~\cite{ptx-mx-mma-frag}.
\begin{equation}
\begin{aligned}
  \small\nonumber
  \textit{mma.m16n8k64.block\_scale.f32.}
  \underbrace{\bstrut{1pt}\textit{e2m1.e2m1}}_{\mathclap{\text{\footnotesize FP4 Precision}}}
  \textit{.f32}
  \underbrace{\bstrut{1pt}\textit{.ue8m0}}_{\mathclap{\text{\footnotesize Scale Format}}}
  \textit{ D, A, B, C, E$_A$, E$_B$}.
  \label{eqn:mma}
\end{aligned}
\end{equation}

\begin{figure}[t]
  \centering
  \includegraphics[width=\columnwidth,clip,trim=0.0in 0.15in 0in 0in]{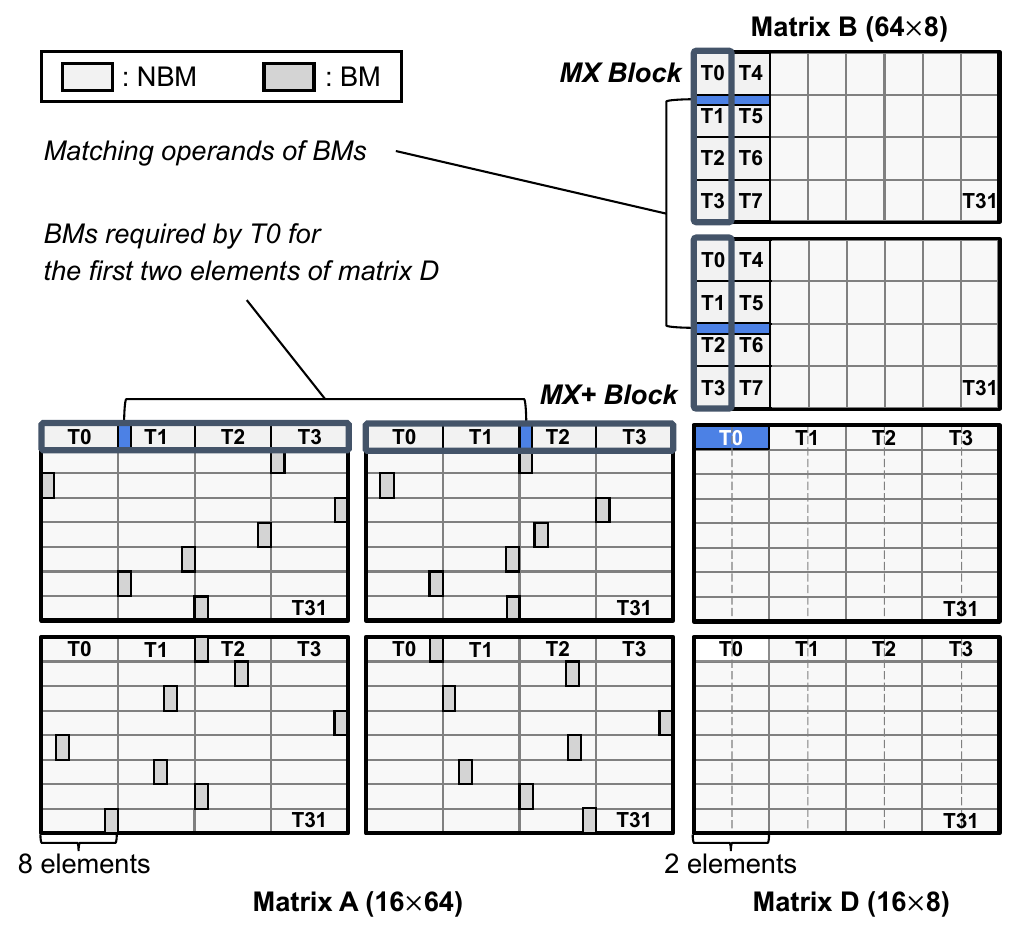}
  \vspace{-0.25in}
  \caption {
            Elements of matrix A (MXFP4+), matrix B (MXFP4), and matrix D
            (FP32) are distributed across the 32 threads (Tx where x is a
            thread ID) in a warp.
           }
  \vspace{-0.15in}
   \label{fig:gpu-bm-handling}
\end{figure}

This instruction operates on matrices A and B with dimensions 16 $\times$ 64
and 64 $\times$ 8, respectively, and matrices C and D with dimensions 16
$\times$ 8. Matrices E$_A$ and E$_B$, each with dimensions 16 $\times$ 2 and 2
$\times$ 8, store the shared exponents of matrices A and B for two MX blocks
per row and per column, respectively.

When a warp executes a machine instruction for an MMA operation, all 32 threads
within the warp collectively perform matrix multiplication for a specific tile
shape. To achieve this, each thread in the warp holds a subset of elements,
referred to as a \emph{fragment}, of the operand matrices in its registers.

\figref{gpu-bm-handling} illustrates how the elements of matrix A (MXFP4+
activation), matrix B (MXFP4 weight), and the resulting matrix D (FP32) are
distributed across the threads in a warp for the 4-bit \textit{mma.m16n8k64}
PTX instruction. 
For matrix A, each thread holds a fragment in four 32-bit registers, with each
register containing eight 4-bit elements, while for matrix B, each thread holds a
fragment in two 32-bit registers, each containing eight 4-bit elements.
For matrix D, each thread holds four 32-bit elements in four 32-bit registers.
Note that each row of matrix A corresponds to two MX+ blocks, while each column
of matrix B represents two MX blocks.

In MXFP4+, BM is effectively represented in E2M3, with the private exponent of
\emax{}, whereas the FP4 compute units in the Tensor Core operate on E2M1.
Thus, we cannot simply perform an MMA operation with the input matrices A and
B. To address this, we decompose the BM value in each MXFP4+ block into the sum
of two values, \bmh{} and \bml{}, as follows:
\begin{equation} 
\small
\begin{aligned}
  BM   &= (-1)^{s} \times 2^{\emax{}} \times um[\text{3:0}] \\
       &= \bmh{} + \bml{}, \\[4pt]
  \textrm{where} ~~ \bmh{} &= (-1)^{s} \times 2^{\emax{}}     \times um[\text{3:2}], \\
                    \bml{} &= (-1)^{s} \times 2^{\emax{} - 2} \times um[\text{1:0}],
  \label{eqn:bm-decomposition}
\end{aligned}
\end{equation}
where $um$[3:0] denotes a mantissa representation with a leading one
\emph{explicitly} stored in $um$[3], as in the x86 80-bit
extended-precision format~\cite{intel-fp}.
As shown in~\eqnref{bm-decomposition}, \bmh{} and \bml{} are effectively E2M1 and can be stored in FP4.
Thus, we can process the BM elements of matrix A using the following steps.
\begin{itemize}
  \item Split BM into \bmh{} and \bml{}.
  \item Replace BM with \bml{} and perform an MMA operation.
  \item Multiply \bmh{} with the corresponding elements in matrix B and accumulate the results into matrix D.
\end{itemize}

Note that additional computation is needed beyond the MMA operation to obtain
the correct output of matrix D (i.e., the third step for \bmh{}).
One possible approach to address this is to exploit CUDA cores with FMA
operations, while the Tensor Core performs an MMA operation for the second
step.
However, we observe that this results in more than a 5$\times$ slowdown in
overall matrix computation with MX+ and MX blocks on the RTX 5090 GPU, compared
to matrix multiplication with only MX blocks.
This is because each FP4 element must be converted to higher precision (e.g.,
BF16 or FP32) to perform FMAs in the CUDA core.
Furthermore, this also requires each thread to fetch the data from other
threads via inter-thread communication (e.g., warp shuffling).
For example, in~\figref{gpu-bm-handling}, thread 0 (\texttt{T0}) needs to fetch
\bmh{}s in matrix A from threads 1 and 2 (\texttt{T1} and \texttt{T2}) and the
matching operands in matrix B from threads 1, 2, 5, and 6 (\texttt{T1},
\texttt{T2}, \texttt{T5}, and \texttt{T6}) to accumulate the multiplication
result into the first two elements of matrix D.

\begin{figure*}[t]
  \includegraphics[width=\textwidth]{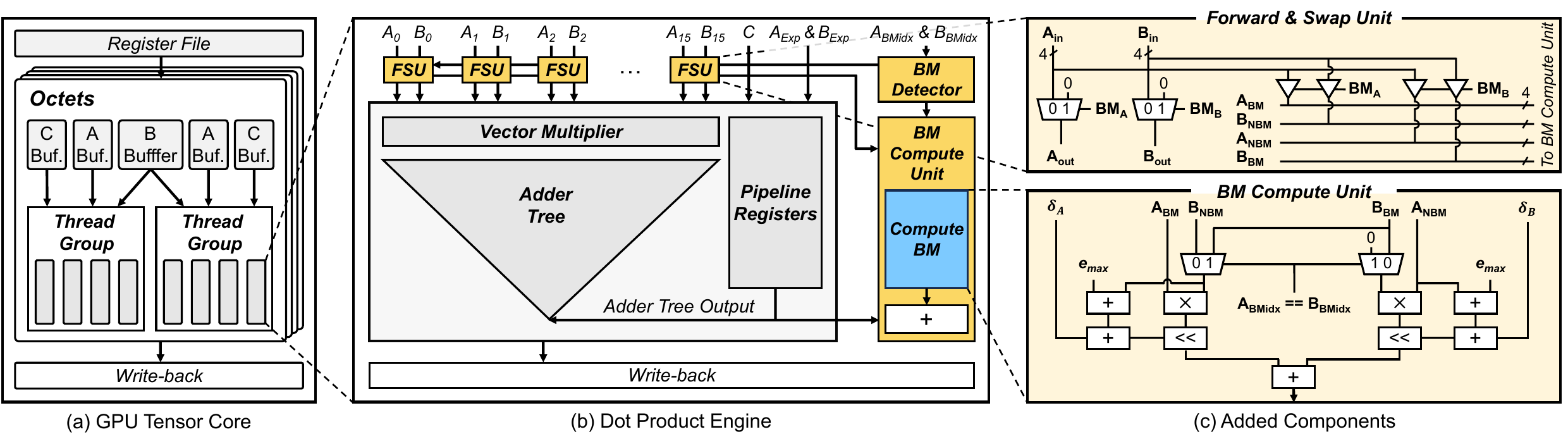}
  \vspace{-0.25in}
  \caption{Overall design of hardware integration of \name{} into GPU.}
  \vspace{-0.10in}
  \label{fig:tensor-core}
\end{figure*}

\putssec{sw-extra-mma}{Using Underutilized Tensor Cores}

To avoid costly conversions and reduce inter-thread communications, we instead
perform an additional MMA operation for \bmh{} while reusing the registers of
each thread.
We observe that this approach maintains performance in the decode stage where
compute units are underutilized, while introducing a modest increase in
inference time during the prefill stage (\ssecref{sw-perf}).

In detail, each thread first loads matrix fragments and the BM index, then
checks if it holds the BM element using its thread ID.
If a thread finds that it holds the BM, it replaces the BM with \bml{}.
For example, in \figref{gpu-bm-handling}, the first four threads
(\texttt{T0}-\texttt{T3}) load the same BM index for the first \name{} block of
matrix A, which is 8, and compare their thread IDs with $\lfloor{\frac{\text{BM
index}}{8}}\rfloor$. 
\texttt{T1} identifies the match and replaces the BM with \bml{}. This process
is repeated across all four registers that hold a fragment of matrix A.

To perform an additional MMA operation using \bmh{} and its matching operands,
each thread requires a fragment of matrix A in separate registers containing
only \bmh{} values, with all other elements set to zero.
To achieve this, we first assign each thread to a single, exclusive \name{}
block, from which it extracts the corresponding \bmh{} from the BM.
For example, in the 16$\times$64 matrix A shown in \figref{gpu-bm-handling},
which contains 32 \name{} blocks, each of the 32 threads in a warp processes a
distinct block and retrieves its \bmh{} value.
Threads then prepare their matrix A fragments in registers for the
additional MMA operation. Those requiring \bmh{} values retrieve them from the
extracting thread and place them at the corresponding BM positions within their
fragments.

\begingroup
\setlength{\textfloatsep}{0pt}
\setlength{\algomargin}{10pt} 
\begin{algorithm}[t]
  \caption{\small Warp executing a sequence of MMA instructions for matrices A
           (MXFP4+) and B (MXFP4).}
  \small
  \label{alg:mmaloop}
  \DontPrintSemicolon
  \SetKwProg{Proc}{procedure}{}{}
  \mycomment{D[mm, nn] += A[mm, 128] $\times$ B[128, nn]}
  \mycomment{Input A, B, and BMIdx are in shared memory}
  \Proc{\normalfont MMALoop(A, B, BMIdx) }{
    \mycomment{Load operands to registers}
    a[$mm$, 128] $\,\gets\,$ LoadFragment(A) \;
    b[$nn$, 128] $\,\,\gets\,$ LoadFragment(B)\;
    bmIdx[$mm$, 4] $\,\gets\,$ Load(BMIdx) \;
    \mycomment{Replace BM of A with \bml{}}
    a $\,\gets\,$ ReplaceBM(a, bmIdx)\;
    \mycomment{Fragments for additional MMAs}
    a$_{BM}$[$mm$, 128] $\,\gets\,$ MakeFragment(A, bmIdx) \;
    \For{\normalfont $k$ \textbf{in} 2}{
      $kk$ $\,\gets$ $k \cdot64$\;
      \For{\normalfont $i$ \textbf{in} $\lceil mm$ / 16$\rceil$}{
        $ii$ $\,\gets\,$ $i \cdot 16$\;
        \For{\normalfont $j$ \textbf{in} $\lceil nn$ / 8$\rceil$}{
          $jj$ $\,\gets\,$ $j \cdot 8$\;
          \textit{mma.m16n8k64} $\,$d[$ii$, $jj$], $\,$a[$ii$, $kk$], $\,$b[$jj$, $kk$], $\,$d[$ii$, $jj$]\;
          \mycomment{Perform additional MMA operation for \bmh{}}
          \If{$k == 1$}{
            \textit{mma.sp.m16n8k128} $\,$d[$ii$, $jj$], $\,$ a$_{BM}$[$ii$], $\,$b[$jj$], $\,$d[$ii$, $jj$]\;
          }
        }
      }
    }
  }
\end{algorithm}
\endgroup

Algorithm~\ref{alg:mmaloop} shows the procedure (\texttt{MMALoop}) for
performing the multiplication of MXFP4+ and MXFP4 matrices.
\texttt{ReplaceBM} (Line~\texttt{9}) identifies the BM in \texttt{a} and
replaces it with \bml{}. 
\texttt{MakeFragment}~(Line \texttt{11}) extracts \bmh{} from the BM and stores
it in the BM position of \texttt{a\textsubscript{BM}}.
Note that this process is amortized over multiple for-loop iterations.
Finally, a sequence of MMA operations is executed, including an additional MMA
(\textit{mma.sp.m16n8k128}) for \bmh{} (Line \texttt{21}), while reusing the
registers that hold fragments of matrices B and D.
We perform a sparse MMA operation, which is twice as fast as a dense MMA,
because all elements in matrix A are zero except \bmh{}.

\putsec{mx-hardware}{Architectural Support for \name{}}

The software integration of \name{} presented in \ssecref{sw-extra-mma} avoids
conversion and reduces inter-thread communication overhead. However, it
requires an additional MMA operation compared to the MX format.
In this section, we explore the hardware integration of \name{} into GPU
systems that support the MX format.

\putssec{}{GPU Integration Overview}

We present a hardware design that supports the \name{} extension without making
intrusive changes to the dot product engine (DPE) in Tensor Cores.
During an MMA operation, Tensor Cores gather the input operands for matrix
multiplication from different threads in a warp.
We capture BMs and their matching operands at the DPE input, then perform
BM-related operations using dedicated low-precision scalar compute units.
This keeps all modifications confined to areas outside the core dot product
pipeline in the DPE.
Since only a few BM-related operands need to be computed while the DPE performs
a dot product, both latency and area overheads remain negligible.
The following section provides a detailed description of this approach.

\putssec{tc-integration}{Tensor Core Integration}

\figref{tensor-core} shows the overall Tensor Core design with \name{} hardware
integration.
Our baseline Tensor Core architecture follows the design in prior
work~\cite{rai:gol19}, with the difference that each warp executes on a single
Tensor Core containing 32 DPEs.
Four threads form a \textit{threadgroup} that utilizes four DPEs, and two
threadgroups combine to form an \textit{octet}. A warp consists of four octets,
with threads collaboratively loading operand matrices into intermediate
buffers. 

Each Tensor Core completes one FP4 \textit{mma.m16n8k64} every 16 cycles
according to our benchmarking, which can also be inferred from the RTX 5090
specification~\cite{RTX5090}.
Since each thread in a warp computes dot products for eight pairs of MXFP4
blocks to produce four output elements during MMA execution
(see~\figref{gpu-bm-handling}), we configure each DPE to process one MXFP4
block pair every two cycles; i.e., each DPE processes 16 FP4 input pairs per
cycle.
Each pair of MXFP6 or MXFP8 blocks is computed every four cycles, since FP8
sustains half the throughput of FP4, and FP6 matches FP8 throughput.
The BM indices of the blocks currently being processed (A\textsubscript{BMidx}
and B\textsubscript{BMidx}) are supplied accordingly.

\myparagraph{Hardware Extension.}
The DPE is extended with three main components: 1) BM Detector, 2) Forward and
Swap Unit (FSU), and 3) BM Compute Unit (BCU).
When the MX+ blocks and BM indices are fed into the DPE, the BM Detector checks
the BM indices (A\textsubscript{BMidx} and B\textsubscript{BMidx}) and
activates the corresponding FSUs by sending them 1-bit BM\textsubscript{A} and
BM\textsubscript{B} signals. 
Each FSU consists of several multiplexers and tri-state buffers, and shares a
datapath connected to the BCU. 
When BM (BM\textsubscript{A}, BM\textsubscript{B}) signals are set, the FSU
directs the BM input and its matching operand to the BCU while forwarding zero
to the corresponding DPE input. This ensures that these inputs are excluded
from computation in the dot product pipeline. 
To support FP6 and FP8, we can configure the FSUs such that those at even
positions (2i, where i = 0, 1, ..., 7) share one datapath, while those at odd
positions (2i+1) share another. This enables the 4-bit inputs from the adjacent
FSU to be routed to the BCU as well.
The forwarded inputs are then processed in the BCU, as discussed below.

\myparagraph{BM Computation within Tensor Core.}
As shown in \figref{tensor-core}(c), the BCU takes as inputs the BMs and their
matching operands along with the BM indices. It then performs the following
computation:
\begin{equation} 
\begin{aligned}
  \nonumber
  Output = (A_{BM} \times B_{NBM})
         + (B_{BM} \times A_{NBM}),
  \label{eqn:bcu}
\end{aligned}
\end{equation}
\noindent where $A_{BM}$ and $B_{BM}$ are the BMs of matrices A and B, and
$B_{NBM}$ and $A_{NBM}$ are their matching operands.
The first and second multiplication terms are conditionally left-shifted by
$\delta_A$ and $\delta_B$ in MX++, which are the differences of the shared
exponents between MX and MX++ for matrices A and B. These shifts are encoded in
the reserved 3 bits of the BM index.
Note that these operations complete faster than the DPE, which performs
element-wise multiplications and multi-level additions using an adder tree, and
do not cause pipeline stalls; i.e., the BM computation does not affect MMA
instruction throughput. 
$Output$ is then added to the output of the adder tree before being normalized
and converted to FP32.

When the BM indices of A and B are identical, we simply swap $B_{NBM}$ with
$B_{BM}$ and set $B_{NBM}$ to zero, effectively computing only one of
the two identical terms in the formula.
We design the multipliers to support sufficiently high precision to handle
scenarios where both the multiplicand and multiplier are BMs.

\begin{figure}[t]
  \includegraphics[width=\columnwidth,clip,trim=0in 0in 0.25in 0in]{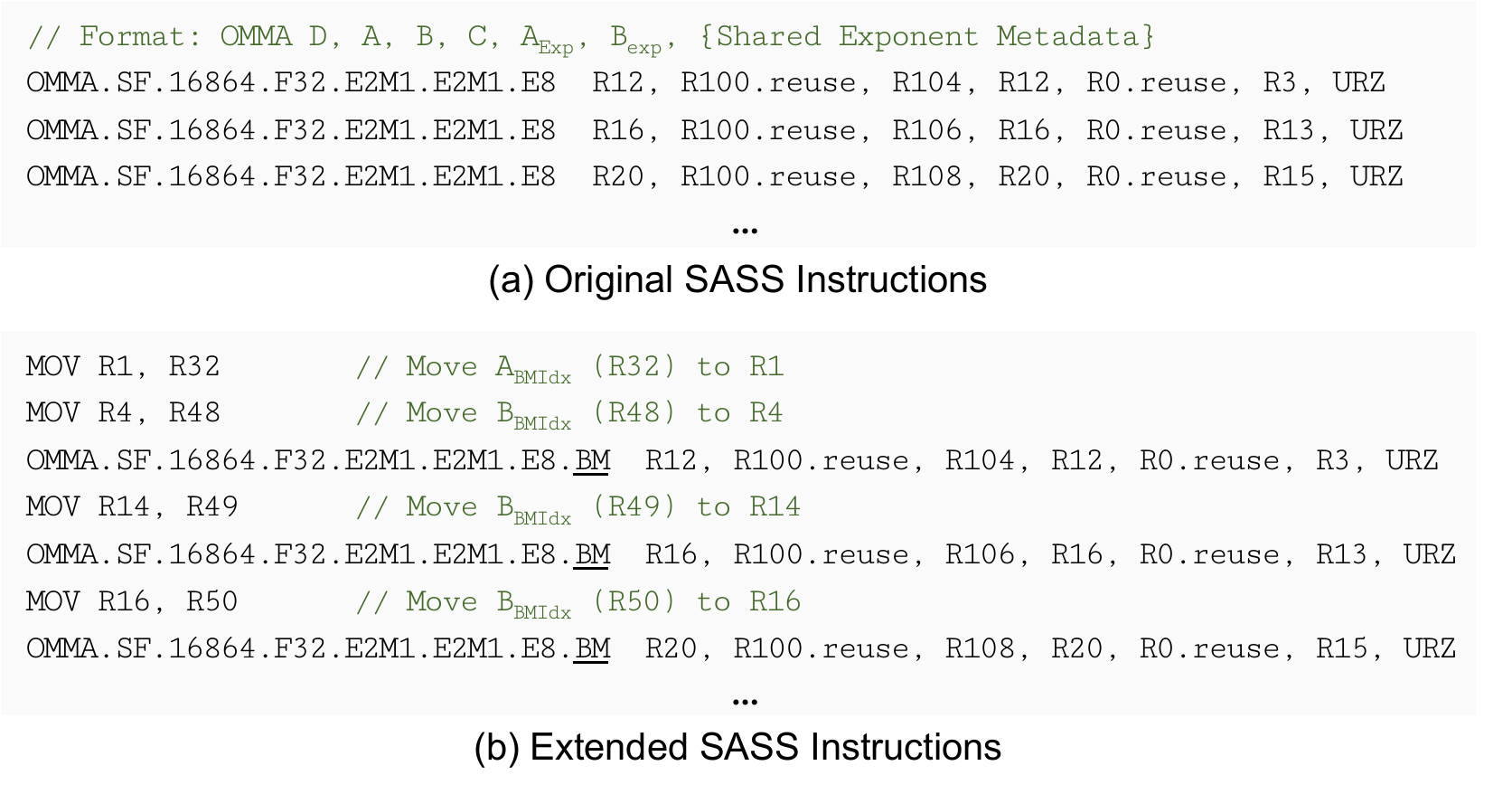}
  \vspace{-0.25in}
  \caption{
           (a) Original SASS instructions performing matrix multiplication.
           (b) Extended SASS instructions with BM flag and additional source
           registers for BM indices.
          }
  \vspace{-0.15in}
  \label{fig:sass}
\end{figure}

\myparagraph{SASS Instruction Extension.}
\figref{sass}(a) shows SASS instructions performing an MXFP4 MMA operation,
corresponding to the MXFP4 $mma$ PTX instruction described in
\ssecref{gpu-challenge}. 
The register identifier in the MMA instruction represents multiple consecutive
registers~\cite{rai:gol19}, with only the lowest register identifier encoded in
the instruction.
For example, \texttt{R12} in the first \texttt{OMMA} instruction represents a
sequence of four registers: \texttt{<R12, R13, R14, R15>}. Note that
\texttt{R0} and \texttt{R3}, which contain the shared exponents, are single
registers rather than register sequences.

\figref{sass}(b) shows the proposed SASS instruction.
The MMA instruction is extended with a \texttt{BM} control flag, which
indicates whether the input operands are represented in \name{}.
We extend the control information by 1 bit for the additional flag; the SASS
instruction contains unused bits~\cite{jia:mag19}, which we similarly observed
for Blackwell instructions through \texttt{nvdisasm}~\cite{nvdisasm}. 
We also extend the MMA instruction to take two additional source registers,
each containing two 8-bit BM indices for matrices A and B, following the
register layout of the shared exponent.

For instruction encoding of BM indices, we follow the scheme used by sparse MMA
instructions for the MX format. 
These instructions implicitly encode the register that holds ordered metadata
together with the register for the shared exponent of matrix A.
The two registers are paired to form a sequence, and a single register
identifier is encoded in the instruction.
We adopt the same approach by pairing the BM index register with the shared
exponent register.
As illustrated in \figref{sass}(b), \texttt{R32} and \texttt{R48}, each
containing BM indices for matrices A and B, are copied to registers \texttt{R1}
and \texttt{R4}, respectively. 
In the subsequent \texttt{OMMA} instruction, \texttt{R0} and \texttt{R3}
implicitly represent the register pairs \texttt{<R0, R1>} and \texttt{<R3,
R4>}.

\putsec{eval}{Evaluation}

\putssec{method}{Experimental Methodology}

\myparagraph{Algorithmic Implementation.}
We implement \name{} on top of the CUDA extension of the MX PyTorch emulation
library~\cite{mx-git} and evaluate model performance using pre-trained models
from Hugging Face~\cite{wol:deb19}. 
Following prior work on applying MX formats to DNNs~\cite{mx-arxiv}, we apply
MX and \name{} formats to all tensors involved in \emph{any} dot product
operation during LLM inference, including those in the language modeling head
and KV cache. We use BF16 for vector operations such as normalization.

\myparagraph{Model and Workload.}
We evaluate \name{} across a diverse set of large language models:
OPT-66B~\cite{zha:rol22}, Llama-3.1 (8B and 70B)~\cite{abh:abh24},
Mistral-7B-v0.3~\cite{jia:sab23}, Phi-4-14B~\cite{abd:ane24}, and
Qwen-2.5-14B-Instruct~\cite{yan:yan25}.
We measure task-specific accuracy (\%) on the same lm-evaluation-harness tasks
used in prior work~\cite{mx-arxiv} and assess language modeling performance
using perplexity on the WikiText-2~\cite{mer:xio16} and C4~\cite{dod:sap21}
datasets.
Additionally, we use Llama-2 models~\cite{tou:mar23} of varying sizes to
measure inference time under \name{} integration.

\myparagraph{Quantization Baselines.}
For model quantization, two primary approaches exist: post-training
quantization (PTQ) and quantization-aware training (QAT). 
Our evaluation on LLMs uses quantized inference based on PTQ, which does not
involve any re-training or fine-tuning.
We follow the drop-in replacement PTQ scenario described in prior
work~\cite{smx}, where a pre-trained BF16 model is directly cast into MX or
\name{} formats for evaluation---referred to as \emph{direct-cast} inference.
To demonstrate that \name{} can also help in other contexts, we additionally
present the results with quantization-aware fine-tuning on vision and CNN
models in~\ssecref{broad-app}.

We assess model performance when using MX+ for both activations and weights,
and compare each against its counterpart (i.e., MXFP4, MXFP6, and MXFP8). We
use E2M3 and E4M3 for six-bit and eight-bit formats, as prior work shows better
accuracy with higher mantissa bit configurations among the two variants in each
format~\cite{mx-arxiv}.
We also evaluate MXFP4++ (applied to both weights and activations) and
\amxfp{} (MXFP4+ only for activations).

\begin{table}[t]
  \caption{
           Direct-cast inference results of zero-shot tasks. Higher is better
           for all tasks. All results are zero-shot except for $\dagger$-marked
           tasks in OPT-66B, which are 5-shot due to poor BF16 performance.
          }
  \vspace{-0.10in}
  \centering
  \resizebox{1.00\columnwidth}{!}{%
    {\fontsize{10}{12}\selectfont
    \renewcommand{\arraystretch}{0.90}
      \begin{tabular}{clcccccc}
        \toprule[1pt]
        \multirow{2}{*}{\textbf{Model}}  &\multirow{2}{*}{\textbf{Format}}
                                         &{\textbf{ARC}}
                                         &{\textbf{ARC}}
                                         &{\textbf{Lam-}}
                                         &{\textbf{College}}
                                         &{\textbf{Int.}}
                                         &{\textbf{Juris-}} \\
                                         &                         
                                         &{\textbf{easy}}
                                         &{\textbf{challenge}}
                                         &{\textbf{bada}}
                                         &{\textbf{CS} $\dagger$}
                                         &{\textbf{law} $\dagger$}
                                         &{\textbf{prudence} $\dagger$} \\
        \midrule                                                                                           
                                         &BF16      &67.26     &39.76    &73.63    &39.00    &29.75    &25.00  \\
        \cmidrule{2-8}                                                                                            
                                         &MXFP8+    &66.88     &39.51    &73.37    &37.00    &31.40    &31.48  \\
                                         &MXFP8    	&65.99     &37.88    &73.18    &28.00    &29.75    &27.78  \\
        \cmidrule{2-8}                                                                                            
                 OPT                     &MXFP6+    &66.58     &40.10    &73.55    &36.00    &28.10    &24.07  \\
                 66B                     &MXFP6    	&66.08     &38.82    &72.66    &34.00    &31.40    &26.85  \\
        \cmidrule{2-8}                                                                                            
                                         &MXFP4++   &62.54     &35.92    &70.70    &29.00    &33.06    &27.78  \\
                                         &MXFP4+    &62.08     &36.86    &69.26    &25.00    &30.58    &23.15  \\
                                         &\amxfp{}  &60.14     &33.96    &67.82    &26.00    &32.23    &30.56  \\
                                         &MXFP4     &35.23     &24.83    &02.97    &24.00    &19.83    &25.93  \\
        \midrule
                                         &BF16      &81.19     &53.33    &75.39    &54.00    &82.64    &73.15  \\
        \cmidrule{2-8}                                                                                            
                                         &MXFP8+    &81.02     &54.01    &75.14    &53.00    &80.99    &71.30  \\
                                         &MXFP8  	  &79.88     &53.16    &75.22    &43.00    &79.34    &75.00  \\
        \cmidrule{2-8}                                                                                            
                 Llama-3.1               &MXFP6+    &80.72     &53.16    &75.26    &51.00    &79.34    &74.07  \\
                 8B                      &MXFP6  	  &80.22     &52.90    &75.06    &47.00    &79.34    &75.00  \\
        \cmidrule{2-8}                                                                                            
                                         &MXFP4++   &70.71     &46.16    &68.10    &34.00    &70.25    &59.26  \\
                                         &MXFP4+    &70.29     &45.65    &66.83    &38.00    &66.12    &54.63  \\
                                         &\amxfp{}  &68.56     &41.47    &66.83    &41.00    &56.20    &50.00  \\
                                         &MXFP4     &49.41     &29.69    &40.17    &26.00    &23.97    &23.15  \\
        \midrule                                                                                                   
                                         &BF16      &86.49     &64.85    &78.91    &64.00    &89.26    &85.19  \\
        \cmidrule{2-8}                                                                                            
                                         &MXFP8+    &86.78     &64.25    &79.22    &63.00    &89.26    &86.11  \\
                                         &MXFP8  	  &87.04     &64.59    &78.71    &65.00    &89.26    &86.11  \\
        \cmidrule{2-8}                                                                                            
                 Llama-3.1               &MXFP6+    &85.40     &63.91    &78.73    &63.00    &86.78    &85.19  \\
                 70B                     &MXFP6  	  &85.27     &63.14    &78.42    &64.00    &88.43    &85.19  \\
        \cmidrule{2-8}                                                                                            
                                         &MXFP4++   &81.65     &58.11    &72.81    &58.00    &84.30    &84.26  \\
                                         &MXFP4+    &79.17     &54.86    &72.70    &57.00    &87.60    &82.41  \\
                                         &\amxfp{}  &78.11     &53.16    &70.68    &52.00    &84.30    &80.56  \\
                                         &MXFP4     &68.18     &44.88    &61.91    &45.00    &75.21    &63.89  \\
        \midrule                                                                                                           
                                         &BF16      &78.32     &52.22    &75.26    &53.00    &76.03    &70.37  \\
        \cmidrule{2-8}                                                                                            
                                         &MXFP8+    &77.90     &51.19    &75.04    &52.00    &73.55    &72.22  \\
                                         &MXFP8  	  &78.54     &52.22    &74.99    &50.00    &74.38    &71.30  \\
        \cmidrule{2-8}                                                                                            
                 Mistral                 &MXFP6+    &78.45     &51.88    &74.85    &51.00    &76.03    &70.37  \\
                 7B                      &MXFP6  	  &78.32     &52.73    &74.93    &52.00    &76.03    &68.52  \\
        \cmidrule{2-8}                                                                                            
                                         &MXFP4++   &75.67     &49.06    &71.43    &42.00    &68.60    &57.41  \\
                                         &MXFP4+    &74.20     &47.78    &70.79    &45.00    &67.77    &65.74  \\
                                         &\amxfp{}  &74.28     &47.78    &72.40    &40.00    &60.33    &54.63  \\
                                         &MXFP4     &69.57     &43.26    &65.17    &31.00    &47.93    &43.52  \\
        \midrule                                                                                      
                                         &BF16      &72.90     &55.97    &72.50    &65.00    &90.91    &83.33  \\
        \cmidrule{2-8}                                                                                            
                                         &MXFP8+    &72.94     &56.23    &72.13    &67.00    &90.08    &81.48  \\
                                         &MXFP8  	  &73.36     &56.74    &72.13    &66.00    &90.08    &84.26  \\
        \cmidrule{2-8}                                                                                            
                 Phi-4                   &MXFP6+    &71.63     &55.46    &71.82    &67.00    &89.26    &82.41  \\
                 14B                     &MXFP6  	  &72.26     &55.46    &71.78    &68.00    &90.08    &83.33  \\
        \cmidrule{2-8}                                                                                            
                                         &MXFP4++   &71.63     &55.46    &69.88    &63.00    &90.08    &78.70  \\
                                         &MXFP4+    &72.47     &54.95    &67.94    &64.00    &90.08    &82.41  \\
                                         &\amxfp{}  &72.31     &54.95    &68.87    &65.00    &88.43    &83.33  \\
                                         &MXFP4     &72.35     &53.24    &64.43    &58.00    &86.78    &84.26  \\
        \midrule                                                                                      
                                         &BF16      &81.52     &62.46    &72.87    &71.00    &87.60    &87.04  \\
        \cmidrule{2-8}                                                                                            
                                         &MXFP8+    &81.27     &61.77    &72.85    &72.00    &88.43    &87.04  \\
                                         &MXFP8  	  &80.81     &61.69    &72.48    &72.00    &88.43    &87.04  \\
        \cmidrule{2-8}                                                                                            
                 Qwen-2.5                &MXFP6+    &81.06     &60.58    &72.02    &71.00    &89.26    &86.11  \\
                 14B                     &MXFP6  	  &80.22     &60.75    &72.23    &70.00    &89.26    &85.19  \\
        \cmidrule{2-8}                                                                                            
                                         &MXFP4++   &78.91     &57.68    &67.88    &67.00    &88.43    &85.19  \\
                                         &MXFP4+    &77.15     &54.61    &66.12    &66.00    &87.60    &82.41  \\
                                         &\amxfp{}  &75.72     &52.90    &65.42    &69.00    &81.82    &82.41  \\
                                         &MXFP4     &69.57     &48.89    &51.89    &48.00    &71.07    &68.52  \\
        \bottomrule[1pt]
      \end{tabular}
    }
  }
  \vspace{-0.10in}
  \label{tab:acc}
\end{table}

\myparagraph{\name{} Software Integration.}
We consider two use-case scenarios for \name{} software integration, based on
whether the system natively supports MX formats: 1) data stored in MX formats
are converted to BF16 before computation, and 2) MX formats are computed directly within hardware.
For the first scenario, we extend the Triton compiler~\cite{til:kun19}---which
already supports MX formats for matrix multiplication via BF16 conversion---by
implementing conversion code for Block Max (BM)
(i.e.,~\eqnref{deconversion-kernel}) within the matrix multiplication kernel. 
We evaluate this using BF16 for activations and MXFP4 variants for weights on
an NVIDIA RTX A6000 GPU~\cite{A6000}, which lacks native MX support.
For the second scenario, we implement our algorithm in \ssecref{sw-extra-mma}
using the CUTLASS library~\cite{cutlass} and integrate our matrix
multiplication kernel into vLLM~\cite{kwo:li23}.
We then evaluate its performance on an NVIDIA RTX 5090 GPU~\cite{RTX5090},
which provides native hardware support for MX formats.

\myparagraph{\name{} Hardware Integration.}
We implement the components added for the \name{} GPU integration in RTL and
synthesize them using Synopsys Design Compiler with a commercial 28nm
technology node.
To evaluate performance with \name{} integration, we extend
AccelSim~\cite{kha:she20} with the configurations similar to the NVIDIA RTX
5090 GPU.
Each instruction for an MMA operation is modified to include additional access
to the register file. We also model the latency of adding the BCU output to the
adder tree result for the MMA instruction.
The matrix multiplication traces are generated using the CUTLASS library. 

\begin{table}[t]
  \caption{
           Perplexity of different models via direct-cast inference. The
           sequence lengths are 1024 (Top) and 2048 (Bottom). Lower is better.
          }
  \vspace{-0.10in}
  \centering
  \resizebox{1.00\columnwidth}{!}{%
    {\fontsize{20}{22}\selectfont
    \renewcommand{\arraystretch}{1.25}
      \begin{tabular}{l|cc|cccc|cc|cc|cc}
        \toprule[2pt]
        \multirow{2}{*}{\textbf{Model}}  &\multicolumn{2}{c|}{\textbf{OPT}}      &\multicolumn{4}{c|}{\textbf{Llama-3.1}} 
                                         &\multicolumn{2}{c|}{\textbf{Mistral}}  &\multicolumn{2}{c|}{\textbf{Phi-4}}  
                                         &\multicolumn{2}{c}{\textbf{Qwen-2.5}}                                          \\[-4pt]
                                         &\multicolumn{2}{c|}{\textbf{66B}}      &\multicolumn{2}{c}{\textbf{8B}} 
                                         &\multicolumn{2}{c|}{\textbf{70B}}      &\multicolumn{2}{c|}{\textbf{7B}}  
                                         &\multicolumn{2}{c|}{\textbf{14B}}      &\multicolumn{2}{c}{\textbf{14B}}       \\
        \midrule
   
        Dataset                          &Wiki2      &C4      &Wiki2      &C4      &Wiki2      &C4       
                                         &Wiki2      &C4      &Wiki2      &C4      &Wiki2      &C4                       \\
        \midrule
        BF16                             &10.66      &10.90   &6.97       &9.40    &3.17       &7.22     
                                         &5.93       &8.26    &7.49       &14.82   &6.46       &10.23                    \\
        \midrule                                                                                                                                         
        MXFP8+                           &10.70      &10.92   &7.06       &9.51    &3.23       &7.26     
                                         &5.95       &8.28    &7.55       &14.89   &6.53       &10.29                    \\        
        MXFP8                            &11.20      &11.11   &7.13       &9.63    &3.28       &7.30                      
                                         &6.00       &8.32    &7.57       &14.91   &6.65       &10.38                    \\
        \midrule                                                                                                                                                          
        MXFP6+                           &10.75      &10.94   &7.10       &9.58    &3.36       &7.34                      
                                         &5.98       &8.30    &7.56       &14.92   &6.61       &10.36                    \\        
        MXFP6                            &11.24      &11.19   &7.19       &9.70    &3.43       &7.39                      
                                         &6.01       &8.33    &7.58       &14.94   &6.72       &10.44                    \\
        \midrule                                                                                                                                         
        MXFP4++                          &12.53     &11.85    & 9.87      &13.08   &5.34       &8.94     
                                         &6.57      &8.97     &8.28       &15.93   &7.98       &11.71                    \\
        MXFP4+                           &12.74     &12.01    &10.14      &13.37   &5.79       &9.36                       
                                         &6.67      &9.08     &8.43       &16.19   &8.31       &12.19                    \\ 
        \amxfp{}                         &13.92     &13.64    &11.03      &14.64   &6.03       &9.65                       
                                         &6.93      &9.45     &8.52       &16.36   &8.85       &12.69                    \\
        MXFP4                            &167.61    &276.80   &27.69      &33.80   &9.15       &13.72                      
                                         &10.06     &13.18    &9.47       &17.45   &13.89      &18.25                    \\
        \bottomrule[2pt] \addlinespace[5pt]
        \toprule[2pt]
        BF16                             &9.35      &10.15    &6.27       &8.62    &2.81       &6.44    
                                         &5.32      &7.81     &6.67       &13.45   &5.70       &9.55                     \\
        \midrule                                                                                                                                          
        MXFP8+                           &9.39      &10.17    &6.35       &8.73    &2.86       &6.48                      
                                         &5.34      &7.83     &6.72       &13.51   &5.76       &9.60                     \\
        MXFP8                            &9.82      &10.38    &6.42       &8.84    &2.91       &6.51                      
                                         &5.36      &7.86     &6.74       &13.53   &5.88       &9.69                     \\
        \midrule                                                                                                                                                         
        MXFP6+                           &9.43      &10.19    &6.38       &8.79    &2.98       &6.56                      
                                         &5.35      &7.85     &6.74       &13.54   &5.83       &9.66                     \\
        MXFP6                            &9.94      &10.44    &6.46       &8.90    &3.04       &6.60                      
                                         &5.38      &7.88     &6.75       &13.56   &5.93       &9.74                     \\
        \midrule                                                                                                                                                           
        MXFP4++                          &11.17     &11.18    &9.22       &13.10   &4.81       &8.34                      
                                         &5.90      &8.52     &7.36       &14.49   &7.11       &10.96                    \\
        MXFP4+                           &11.35     &11.31    &9.54       &13.41   &5.25       &8.58                      
                                         &5.97      &8.62     &7.48       &14.73   &7.38       &11.42                    \\
        \amxfp{}                         &12.63     &13.63    &10.46      &15.00   &5.45       &8.96                      
                                         &6.27      &9.04     &7.58       &14.87   &7.92       &11.93                    \\
        MXFP4                            &209.83    &306.33   &27.38      &36.41   &8.43       &13.80                      
                                         &9.96      &14.40    &8.45       &15.99   &12.28      &17.46                    \\
        \bottomrule[2pt]
      \end{tabular}
    }
  }
  \vspace{-0.20in}
  \label{tab:ppl}
\end{table}

\putssec{model-perf}{Language Model Performance}

\tabref{acc} shows the accuracy of the baseline BF16, MX, and \name{} for the
lm-evaluation-harness tasks used in~\cite{mx-arxiv}. 
Overall, \name{} improves accuracy over its MX counterparts, with MXFP8+ and
MXFP6+ achieving improvements of up to 10.00 and 4.00 percentage points.
The accuracy difference between MXFP4 and MXFP4+ is particularly substantial
(up to +42.15\%, excluding OPT-66B where MXFP4 does not work), and even the
case where MXFP4+ is used only for activations (A-MXFP4+), while MXFP4 is still
used for weights, still significantly outperforms MXFP4. 
This indicates that representing activation outliers becomes challenging in
low-precision formats, and \name{} effectively addresses this problem by
representing BMs with higher precision.
Importantly, MXFP4+ achieves this while maintaining hardware and memory
efficiency by representing all elements uniformly in four bits like MXFP4.
Building on MXFP4+, MXFP4++ further improves accuracy by also representing NBMs
more accurately, achieving up to +4.63\% higher accuracy compared to MXFP4+.
Consistent with the accuracy results, \tabref{ppl} shows that both \name{}
and MX++ \textit{always} achieve lower perplexity than the original MX formats 
across sequence lengths and datasets.

\begin{table}[t]
   \caption{
            Matrix multiplication time with BF16 activation and MXFP4+ or
            MXFP4++ weight, normalized to the MXFP4 weight case.
           }
  \vspace{-0.10in}
  \centering
  \resizebox{1.00\columnwidth}{!}{%
  {\fontsize{10}{12}\selectfont
    \renewcommand{\arraystretch}{1.05}
      \begin{tabular}{l|ccc|ccc}
        \toprule[0.8pt]
        Normalized Time    &\multicolumn{3}{c|}{Small Activations} 
                           &\multicolumn{3}{c}{Large Activations}                           \\
        N=4096, K=4096     &M=8      &M=16      &M=32      &M=1024     &M=2048    &M=4096   \\
        \midrule[0.4pt]                                   
        MXFP4+             &1.08     &1.07      &1.08      &1.04       &1.01      &1.01     \\
        MXFP4++            &1.08     &1.09      &1.10      &1.04       &1.05      &1.04     \\
        \bottomrule[0.8pt]
      \end{tabular}
    }
  }
  \label{tab:triton}
  \vspace{-0.15in}
\end{table}

\putssec{sw-perf}{Performance of \name{} Software Integration}

In this section, we evaluate two use-case scenarios of \name{} software
integration, as discussed in~\ssecref{method}.

\myparagraph{Conversion Before Computation.}
\tabref{triton} presents the execution time for matrix multiplication using
BF16 activations and MXFP4+ (or MXFP4++) weights, normalized to the MXFP4
weight case.
The execution times reported in~\tabref{triton} include both BF16 conversion
overhead and BF16 MMA operations. Note that no additional MMA operation is
required for MXFP4+ (or MXFP4++) in this case.
Matrix multiplication kernels are generated from Triton~\cite{til:kun19} for
the dimensions spanning from low (small activations) to high (large
activations) data reuse scenarios.
The results show that the BM handling overhead during BF16 conversion becomes
more pronounced with smaller activations than larger ones.
In high reuse cases, BF16 MMAs dominate the overall matrix multiplication time,
while amortizing the conversion overhead. Note that, in both cases, the
additional BM handling required during conversion introduces only a small
performance overhead over the MX.

\begin{figure}[t]
  \includegraphics[width=1.00\columnwidth,clip,trim=0.0in 0.15in 0in 0in]{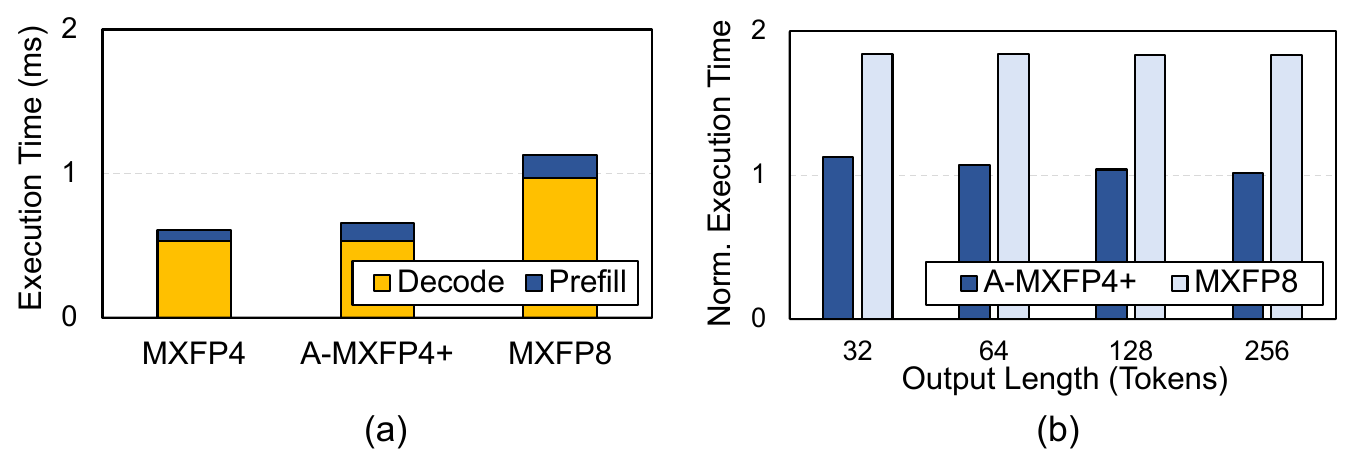}
  \vspace{-0.25in}
  \caption{ 
           (a) Execution time breakdown with 64 output tokens.
           (b) Normalized execution time across output tokens.
          }
  \vspace{-0.15in}
  \label{fig:gpu-sw-perf}
\end{figure}

\myparagraph{Direct Computation.}
In the following sections, we refer to the aggregated matrix multiplication
time during LLM inference in vLLM for a given number of concurrent requests as
\textit{execution time}.
\figref{gpu-sw-perf}(a) shows the execution time during the prefill and decode
stages for Llama-2-13B, with four requests of 1024 input tokens and 64 output
tokens.
The results show that \amxfp{} performs close to MXFP4, while MXFP8 leads to a
large slowdown.
Since the decode stage that dominates the execution time is memory-bounded,
an additional MMA operation in \amxfp{} incurs a negligible performance overhead
(6.71\%).
{\amxfp{} shows a moderate slowdown in the prefill stage (1.54$\times$), which
comprises 18.78\% of the execution time.}
\figref{gpu-sw-perf}(b) shows the execution time across different output
tokens, normalized to MXFP4. 
\amxfp{} shows up to a 1.13$\times$ slowdown, while MXFP8 shows up to a
1.85$\times$ slowdown compared to MXFP4.
As the number of output tokens increases in \figref{gpu-sw-perf}(b), the
decode stage accounts for a larger portion of the execution time, reducing the
gap between MXFP4 and \amxfp{}. 

\begin{figure}[t]
  \includegraphics[width=1.00\columnwidth,clip,trim=0.0in 0.00in 0in 0in]{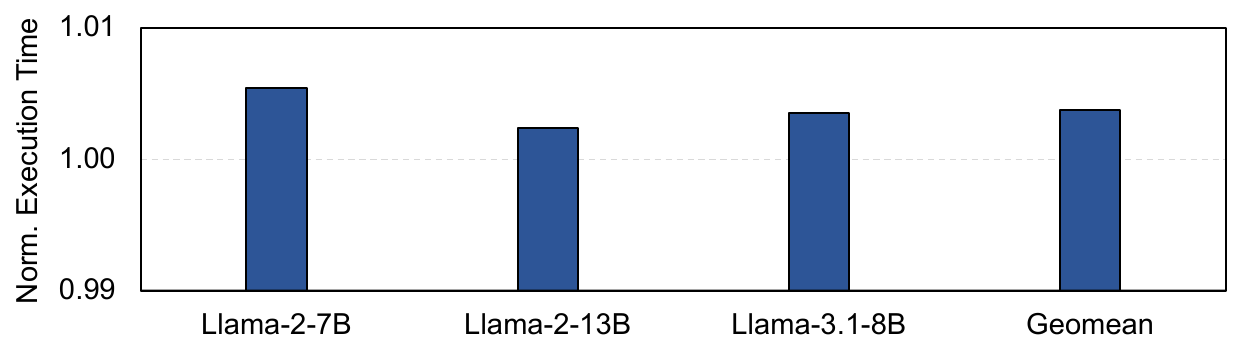}
  \vspace{-0.25in}
  \caption{Normalized execution time of MXFP4+.}
  \vspace{-0.05in}
  \label{fig:hw-perf}
\end{figure}

\putssec{}{\name{} Hardware Integration}

\myparagraph{Performance.}
\figref{hw-perf} shows the execution time of MXFP4+ with hardware integration
during the prefill stage for a request with 2048 input tokens, normalized to
MXFP4.
Overall, MXFP4+ shows a 0.38\% slowdown on average compared to MXFP4.
This is because the BCU computation does not affect the throughput of the
instructions for MMA operations.
Extra register file access and increased instruction latency have a
negligible impact on performance.

\begin{table}[b]
  \vspace{-0.10in}
  \caption{Area and power for \name{} support per Tensor Core.}
  \vspace{-0.10in}
  \centering
  \resizebox{0.95\columnwidth}{!}{%
    {\fontsize{10}{12}\selectfont
      \begin{tabular}{llrr}
        \toprule[0.8pt]
        Component                   & Configuration           & Area [mm$^2$]  & Power [mW]    \\
        \midrule[0.4pt]
        Forward and Swap Unit       & 32 $\times$ (16 units)  & 0.004          & 0.59          \\
        BM Detector                 & 32                      & 0.004          & 2.86          \\
        BM Compute Unit             & 32                      & 0.012          & 8.66          \\
        \midrule[0.4pt]
        \textbf{Total}              &                         &\textbf{0.020}  &\textbf{12.11} \\
        \bottomrule[0.8pt]
      \end{tabular}
    }
  }
  \label{tab:area-power}
\end{table}

\myparagraph{Area and Power.}
\tabref{area-power} shows the area and power of the additional components for
\name{} per Tensor Core. 
We add 16 FSUs, a BM Detector, and a Compute Unit for each 32 DPEs in the
Tensor Core.
Our design has an area of {0.020}mm$^2$ and a power consumption of 12.11mW.
Note that directly comparing the area with the RTX 5090 is not feasible since
we use a 28nm technology node while the GPU is fabricated using a more advanced
node (4nm). However, we believe the area overhead would be even smaller if
fabricated using more advanced node. 
The area overhead of \name{} is much smaller than recent quantization work that
integrates hardware components into Tensor Cores, such as
RM-STC~\cite{hua:wan23} and OliVe~\cite{guo:tan23}.

\begin{figure}[t]
  \includegraphics[width=1.00\columnwidth]{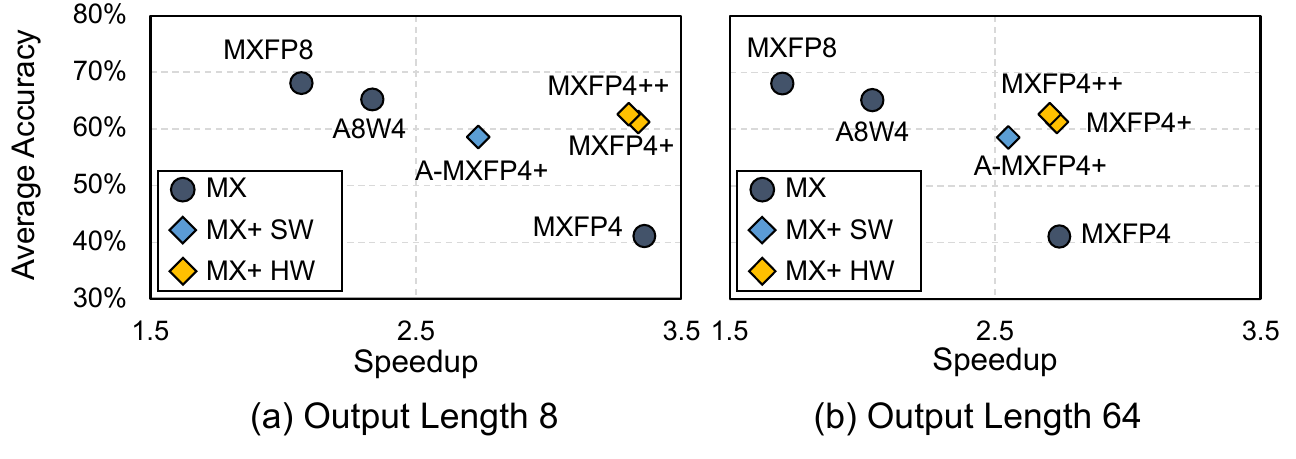}
  \vspace{-0.25in}
  \caption{
           End-to-end inference speedup over BF16 and average accuracy of
           lm-eval-harness tasks on Llama-2-13B. A8W4 uses MXFP8 for
           activations and MXFP4 for weights.
          }
  \vspace{-0.10in}
  \label{fig:e2e}
\end{figure}

\myparagraph{End-to-End Speedup.}
\figref{e2e} shows the speedup over BF16 in vLLM and the average accuracy for
lm-eval-harness tasks on Llama-2-13B. We evaluate performance using four
requests of 1024 input tokens with either 8 or 64 output tokens, representing
scenarios where the prefill or decode stage respectively dominates inference
time.
We compare MX formats with MXFP4+ and MXFP4++ under hardware support, as well as
\amxfp{} under software support. 
For the long output length, \amxfp{} achieves a speedup close to MXFP4 with
17.46\% higher accuracy. 
We modify the CUTLASS library to reduce unnecessary data loading and
computation when an output matrix dimension is smaller than
the tile size of a thread block. The library supports a single tile shape for A8W4,
where each thread block computes an M=128 and N=128 output tile~\cite{cutlass-tile-size}, whereas
M of the output matrix is {4} in the decode stage. Even with this optimization, A8W4
performance remains close to MXFP8. 

\begin{table}[t]
  \caption {Total quantization time normalized to MXFP4.}
  \vspace{-0.10in}
  \centering
  \resizebox{0.85\columnwidth}{!}{%
    {\fontsize{5}{5}\selectfont
    \renewcommand{\arraystretch}{0.70}
      \begin{tabular}{lccccc}
        \toprule[0.4pt]
        Input Tokens      &32       &128      &512       &1024       &2048    \\
        \midrule[0.2pt]
        MXFP4+            &1.00     &1.00     &1.03      &1.05      &1.05     \\[1.5pt]
        MXFP4++           &1.05     &1.04     &1.13      &1.15      &1.15     \\
        \bottomrule[0.4pt]
      \end{tabular}
    }
  }
  \label{tab:conversion-time}
  \vspace{-0.10in}
\end{table}

With hardware support, MXFP4+ delivers speedups comparable to MXFP4, achieving
3.34$\times$ and 2.73$\times$ improvements over BF16 in prefill and
decode-dominant scenarios, respectively, while providing 20.17\% higher
accuracy.
Despite requiring additional computation to find the second maximum magnitude
during conversion, MXFP4++ maintains competitive performance, running only
1.04\% and 1.00\% slower than MXFP4. 
\tabref{conversion-time} shows the quantization time across the input token
lengths. 
MXFP4+ exhibits quantization times similar to MXFP4, while MXFP4++ shows a
small increase.
Since quantization accounts for only a small portion of inference time, this
overhead has a negligible impact on overall performance.

\putsec{}{Analysis and Discussion}

\begin{table}[t]
  \caption {Perplexity on WikiText-2 via direct-cast inference.}
  \vspace{-0.10in}
  \centering
  \resizebox{1.00\columnwidth}{!}{%
    {\fontsize{12}{14}\selectfont
    \renewcommand{\arraystretch}{1.10}
      \begin{tabular}{l|c|ccc|cc|c|c|c}
        \toprule[1.1pt]
        \multirow{2}{*}{\textbf{Scheme}}  &\textbf{OPT} 
                                          &\multicolumn{3}{|c|}{\textbf{Llama-2}}
                                          &\multicolumn{2}{|c|}{\textbf{Llama-3.1}}
                                          &\textbf{Mistral} 
                                          &\textbf{Phi-4}  
                                          &\textbf{Qwen-2.5}                        \\
                                          &\textbf{66B}
                                          &\textbf{7B} &\textbf{13B} &\textbf{70B}
                                          &\textbf{8B} &\textbf{70B}
                                          &\textbf{7B} 
                                          &\textbf{14B}
                                          &\textbf{14B}                                \\
        \midrule
        BF16  	                          &9.35          &5.47          &4.89             
                                          &3.32          &6.27          &2.81             
                                          &5.32          &6.67          &5.70          \\
        \midrule
        SMQ (INT4)  	                    &5E+4          &3E+3          &5E+3                   
                                          &1E+3          &8E+3          &2E+4                   
                                          &1E+3          &32.50         &3E+3          \\
        SMQ (MXFP4)  	                    &33.96         &10.44         &9.94                   
                                          &6.03          &21.00         &7.38                   
                                          &9.72          &8.13          &10.13         \\
        QuaRot (INT4)                     &138.88        &8.11          &6.07                   
                                          &4.12          &9.85          &35.76                  
                                          &6.32          &7.55          &9.71          \\
        QuaRot (MXFP4)                    &13.49         &13.41         &7.12                   
                                          &4.17          &9.60          &19.47                  
                                          &6.63          &7.63          &8.17          \\
        Atom (INT4+INT8)                  &\textbf{9.42} &5.94          &5.21                   
                                          &3.66          &7.43          &4.22                   
                                          &5.70          &7.16          &6.77          \\
        \midrule
        ANT           	                  &1E+4          &180.81        &146.11           
                                          &32.83         &927.90        &1E+5             
                                          &4E+2          &22.06         &1E+2          \\
        OliVe         	                  &6E+3          &86.27         &3E+3             
                                          &97.88         &3E+2          &4E+4             
                                          &43.40         &14.66         &13.59         \\
        Tender                            &12.38         &36.47         &55.08                  
                                          &13.43         &70.74         &3E+4                   
                                          &38.88         &19.40         &2E+2          \\
        \midrule
        MX-ANT                            &10.35         &6.10          &5.33                   
                                          &3.79          &8.11          &4.37                   
                                          &5.88          &7.17          &6.81          \\
        MX-OliVe      	                  &20.89         &6.33          &5.50                   
                                          &3.91          &8.31          &4.71                   
                                          &6.01          &7.31          &7.24          \\
        MX-Tender                         &9.58          &6.32          &5.47                   
                                          &3.84          &9.42          &11.58                  
                                          &6.22          &7.25          &7.63          \\
        \midrule
        \textbf{MXFP4+}                   &{10.54}       &\textbf{5.87} &\textbf{5.17} 
                                          &\textbf{3.58} &\textbf{7.22} &\textbf{3.85} 
                                          &\textbf{5.65} &\textbf{6.97} &\textbf{6.66} \\
        \textbf{MXFP4++}                  &{10.46}       &\textbf{5.84} &\textbf{5.16} 
                                          &\textbf{3.56} &\textbf{7.14} &\textbf{3.78} 
                                          &\textbf{5.58} &\textbf{6.95} &\textbf{6.54} \\
        \bottomrule[1.1pt]
      \end{tabular}
    }
  }
  \label{tab:vs_others}
  \vspace{-0.15in}
\end{table}

\putssec{ol-quant}{Comparison with Other Schemes}

In this section, we compare model performance between \name{} and other
algorithm-only or algorithm-hardware quantization schemes. 
For a fair comparison, we quantize matrix multiplication between weights and
activations, excluding language modeling head---the intersection of quantized
operations across the schemes.

SmoothQuant~\cite{xia:lin22} rescales activation channels, while
QuaRot~\cite{ash:moh24} rotates activation using orthogonal matrices to reduce
the overall magnitude. 
Atom~\cite{zha:lin} reorders channels and quantizes the channels with outliers
using higher precision (INT8).
\tabref{vs_others} shows that SmoothQuant (SMQ) falls short in 4-bit precision,
as discussed in multiple studies~\cite{liu:liu23,tender,ash:moh24}.
We observe that QuaRot does not completely remove outliers and performs worse
than MXFP4+; the magnitude of outlier values is not reduced after rotation
(e.g., the down projection layer in Llama-3.1).
\name{} focuses on precisely representing important outlier values, unlike
QuaRot, as the fine-grained grouping of MX already limits the impact of
outliers to other values.
Atom shows comparable model performance as it represents outliers in
8-bit but still performs worse than \name{}.

ANT~\cite{guo:che22} and OliVe~\cite{guo:tan23} use custom formats, while
Tender~\cite{tender} groups channels of similar range and uses standard INT4.
As shown in the table, they suffer at 4-bit due to coarse channel-group or
tensor-wise grouping. 
We extend the schemes to support finer-grained grouping solely for accuracy
comparison purposes, referred to as MX-ANT, MX-OliVe, and MX-Tender, though
this would noticeably increase their runtime overhead.
MX-Tender groups channels of each two rows at runtime.
MX-ANT and MX-OliVe support group-wise quantization of size 32. Both adaptively
select per-group data types for weights and per-tensor data types for
activations. All schemes use floating-point scaling factors calculated per
group at runtime.
Nevertheless, \name{} still shows better performance, demonstrating the
effectiveness of representing BMs in high precision. 

\begin{table}[t]
  \caption{
           Direct-cast perplexity of different weight formats under BF16
           activation with AWQ (A16W4) and MXFP8 activation (A8W4).
          }
  \vspace{-0.10in}
  \centering
  \resizebox{1.00\columnwidth}{!}{%
    {\fontsize{6}{7}\selectfont
    \renewcommand{\arraystretch}{0.95}
      \begin{tabular}{l|ccc|cc}
        \toprule[0.5pt]
        \textbf{Activation}       &\multicolumn{3}{c|}{\textbf{BF16 \& AWQ}} 
                                  &\multicolumn{2}{c}{\textbf{MXFP8}}                \\[1pt]
        \textbf{Weight}           &\textbf{INT4}   &\textbf{MXFP4}  &\textbf{MXFP4+}  
                                  &\textbf{MXFP4}  &\textbf{MXFP4+}                  \\
        \midrule[0.25pt]
        Llama-3.1-8B              &6.67            &6.99            &\textbf{6.64}  
                                  &7.40            &\textbf{6.97}                    \\
        \midrule[0.25pt]                                                                                        
        Mistral-7B                &5.44            &5.56            &\textbf{5.42}  
                                  &5.67            &\textbf{5.52}                    \\
        \bottomrule[0.5pt]
      \end{tabular}
    }
  }
  \label{tab:woq}
  \vspace{-0.10in}
\end{table}

\putssec{broad-app}{Broader Applicability of \name{}}

\myparagraph{Weight-Only Quantization.}
While \name{} primarily targets the scenario where we want to use low-bit
precision for both weight and activation tensors, it also provides the benefit
for weight-focused quantization scenarios.
\tabref{woq} shows the perplexity when employing 4-bit data formats for weights,
with activations in either BF16 using AWQ~\cite{lin:tan23} or MXFP8. 
AWQ is a weight-only quantization method, which scales important weight
channels to larger magnitudes to protect them under low-bit quantization.
Although directly using MXFP4 with AWQ degrades model performance, MX+ can
synergistically operate with AWQ. This is because scaling up the important
channels allows more important weight elements to be identified as BM.
As a result, the model performance is further enhanced compared to the original
AWQ (Weight INT4). 
When using MXFP8 activations with MXFP4 weights, precise representation of
weights can be more critical than activations, as it uses half the number of
bits.
Using MXFP4+ under this setting noticeably enhances the model performance, as
shown in the table.

\begin{table}[t]
  \caption{Top-1 accuracy (\%) on ImageNet.}
  \vspace{-0.10in}
  \centering
  \resizebox{1.00\columnwidth}{!}{%
    {\fontsize{11}{13}\selectfont
    \renewcommand{\arraystretch}{1.15}
      \begin{tabular}{ll|c|cc|cc}
        \toprule[1pt]
        \multirow{2}{*}{\textbf{Family}}  &\multirow{2}{*}{\textbf{Model}} 
                                          &\multirow{2}{*}{\textbf{FP32}}     
                                          &\multicolumn{2}{c|}{\textbf{Direct-cast}}       
                                          &\multicolumn{2}{c}{\textbf{QA fine-tuning}}\\[0.5pt]
                             &            &        &\textbf{MXFP4} &\textbf{MXFP4+} &\textbf{MXFP4} &\textbf{MXFP4+} \\
        \midrule[0.5pt]
        Vision               &DeiT-Tiny   &71.64   &61.79          &66.60           &68.96          &69.96           \\
        Transformer          &DeiT-Small  &79.82   &73.82          &77.29           &76.96          &77.44           \\
        \midrule[0.5pt]
        \multirow{2}{*}{CNN} &ResNet-18   &69.18   &49.28          &62.66           &65.76          &66.72           \\
                             &ResNet-34   &74.55   &52.67          &64.35           &69.97          &71.40           \\
        \bottomrule[1pt]
      \end{tabular}
    }
  }
  \label{tab:vision}
  \vspace{-0.15in}
\end{table}

\myparagraph{Other DNN Workloads.} 
We also evaluate \name{} on Vision Transformer~\cite{tou:cor21} and CNN
models~\cite{he:zha16} for image classification tasks on the ImageNet
dataset~\cite{rus:den15}. 
\tabref{vision} shows that the accuracy of MXFP4+ is higher compared to MXFP4,
with improvements of up to +4.81\% and +13.38\% for the DeiT and ResNet models,
under direct-cast inference. 
As discussed in prior work~\cite{liu:liu23,don:lu23,son:fu20,zha:hu19}, we
observe that activation outliers are also present in these models and are
typically scattered across MX blocks.
\name{} represents these outliers more precisely, thereby improving accuracy.

We also perform quantization-aware (QA) fine-tuning and evaluate the
effectiveness of MXFP4+ on the fine-tuned models.
Compared to direct-cast inference, the accuracy gap between MXFP4 and MXFP4+
becomes narrower, as the fine-tuned models for image classification tasks can
achieve accuracy close to the FP32 baseline even when using MXFP4.
However, for more complex models and challenging tasks where fine-tuned MXFP4
models cannot reach FP32-level accuracy, the difference in accuracy between
MXFP4 and MXFP4+ is likely to be more pronounced.

\begin{table}[t]
  \caption{
           Perplexity on WikiText-2 across non-FP microscaling formats in a
           direct-cast setting.
          }
  \vspace{-0.10in}
  \centering
  \resizebox{\columnwidth}{!}{%
    {\fontsize{5}{6}\selectfont
    \renewcommand{\arraystretch}{0.40}
      \begin{tabular}{l|cc|cc}
        \toprule[0.4pt]
        \textbf{Model}   &\textbf{MXINT8+} &\textbf{MXINT8} &\textbf{MXINT4+} &\textbf{MXINT4} \\
        \midrule[0.2pt]
        Llama-3.1-8B     &6.286            &6.287           &12.850           &14.339          \\
        \midrule[0.2pt]                                                                                                       
        Mistral-7B       &5.321            &5.321           &6.841            &7.156           \\
        \bottomrule[0.4pt]
      \end{tabular}
    }
  }
  \label{tab:non-fp}
  \vspace{-0.10in}
\end{table}

\myparagraph{Applicability of MX+ to Non-FP Microscaling Formats.}
Beyond the three MXFP variants, the OCP MX specification defines one additional
MX-compliant format with integer element data type: MXINT8.
Although MXINT8 lacks an exponent field in its element data type, the approach
of adding extra precision to the BM element could be similarly applied to
MXINT8. 
For instance, the INT8 encoding in MXINT8 uses one sign bit, one integer bit,
and six fractional bits.
In this configuration, \emax{} equals zero in~\eqnref{mxfp-scale} since element
values are always smaller than 2. The shared exponent becomes simply the
exponent of the BM value, while the BM element is represented in the
$\pm$1.xxxxxx format.
This allows us to potentially make the integer bit implicit and use it as an
extra fractional bit for the BM element.
\tabref{non-fp} shows perplexity results for this method applied to MXINT8
and a \emph{hypothetical} MXINT4 format (one sign bit, one integer
bit, two fractional bits). 
For MXINT8, increasing fraction bits from six to seven barely helps. In
contrast, MXINT4 benefits from additional fraction bits, similar to MXFP4+ or
MXFP6+. If MXINT4 becomes part of the concrete MX-compliant formats, this
direction might be worth exploring as well.

\begin{table}[t]
  \caption{
           Direct-cast inference accuracy for NVFP4 and NVFP4+ (NVFP4 with
           extra precision for BM) on lm-eval-harness tasks.
          }
  \vspace{-0.10in}
  \centering
  \resizebox{\columnwidth}{!}{%
    {\fontsize{14}{16}\selectfont
    \renewcommand{\arraystretch}{1.20}
      \begin{tabular}{llcccccc}
        \toprule[1.2pt]
        \textbf{Model}    		         &\textbf{Scheme}  &\multiline{\textbf{ARC}}{\textbf{easy}} 
                                                         &\multiline{\textbf{ARC}}{\textbf{challenge}} 
                                                         &\multiline{\textbf{Lam-}}{\textbf{bada}} 
                                                         &\multiline{\textbf{College}}{\textbf{CS}} 
                                                         &\multiline{\textbf{Int.}}{\textbf{law}} 
                                                         &\multiline{\textbf{Juris-}}{\textbf{prudence}} \\
        \midrule
        \multirow{2}{*}{Llama-3.1-8B}  &NVFP4            &68.81  &45.73  &56.51  &30.00  &63.64  &53.70  \\
        \cmidrule{2-8}                                                                                       
                                       &NVFP4+           &72.31  &46.25  &69.36  &35.00  &71.90  &56.48  \\
        \midrule
        \multirow{2}{*}{Mistral-7B}    &NVFP4            &74.03  &48.46  &70.25  &41.00  &69.42  &60.19  \\
        \cmidrule{2-8}                                                                                       
                                       &NVFP4+           &74.75  &48.63  &70.81  &43.00  &74.38  &62.96  \\
        \bottomrule[1.2pt]
      \end{tabular}
    }
  }
  \label{tab:nvfp4}
  \vspace{-0.10in}
\end{table}

\myparagraph{Applicability of MX+ to NVFP4.}
NVIDIA recently introduced NVFP4, a 4-bit floating-point format that bears
similarity to MXFP4. Both formats use FP4 (E2M1) elements with a shared scale
per block. However, NVFP4 differs by using a smaller block size of 16 elements
and an E4M3 scale factor.
\tabref{nvfp4} presents the direct-cast accuracy results for NVFP4. 
When compared to the results in~\tabref{acc}, MXFP4+ and MXFP4++ perform better
than or comparably to NVFP4. This is because outliers are typically represented
more accurately in \name{} due to the extra precision for BMs.

The \name{} extension can be similarly applied to NVFP4 since both MXFP4 and
NVFP4 map the BM as closely as possible to the maximum representable magnitude
in FP4 when computing scale factors~\cite{nvfp4}.
Similar to \name{}, we extend the mantissa bits of the BM element in NVFP4,
except when the BM magnitude is extremely small that the exponent in the
element data type is not set to maximum (i.e., when the shared scale becomes
$X_{E4M3}\leq00000010_{2}$). 
In such cases, we use the original NVFP4 representation for the block.  
Note that the frequency of such cases can be reduced through an extra
per-tensor software scaling step, which shifts values to larger magnitudes for
per-block scaling.
The extended NVFP4, termed NVFP4+, has an additional 4 bits per block to store
the BM index, which can be packed with BM indices from other blocks for byte
alignment.
As shown in~\tabref{nvfp4}, NVFP4+ achieves higher accuracy than NVFP4.

\myparagraph{Support for MX+ in Systolic Array Variants.} 
Instead of performing matrix multiplications on GPUs, one can also consider
using fixed-function matrix pipelines such as TPU~\cite{jou:you17}.
These pipelines typically implement weight-stationary or output-stationary
systolic array designs~\cite{jou:you17, kap:rom24}, where each processing
element (PE) performs one MAC operation per cycle.
Supporting \name{} in these pipelines can be done similarly to the GPU
integration by adding FSUs and BCUs to the datapath. 
For example, in a representative 32$\times$32 MX-compliant systolic array, an
FSU is attached to each PE, with a single BCU shared among the PEs in each
column.
In the weight-stationary dataflow, the PEs in a column collectively perform a
dot product of an MX block pair. 
The BCU is located below the systolic array and receives BM values and their
matching operands forwarded by FSUs, along with a partial sum.
It then computes the BM-related operands, adds the result to the partial sum,
and forwards the value to the accumulator.
The process is similar for the output-stationary dataflow.
Each PE performs a dot product of an MX block pair over 32 cycles, with FSUs
collecting BM-related operands. After these cycles, the operands and partial
sum are forwarded to the BCU. The updated partial sum is then routed back to
each PE where the accumulator resides.

\begin{figure}[t]
  \includegraphics[width=1.00\columnwidth,clip,trim=0.0in 0.15in 0in 0in]{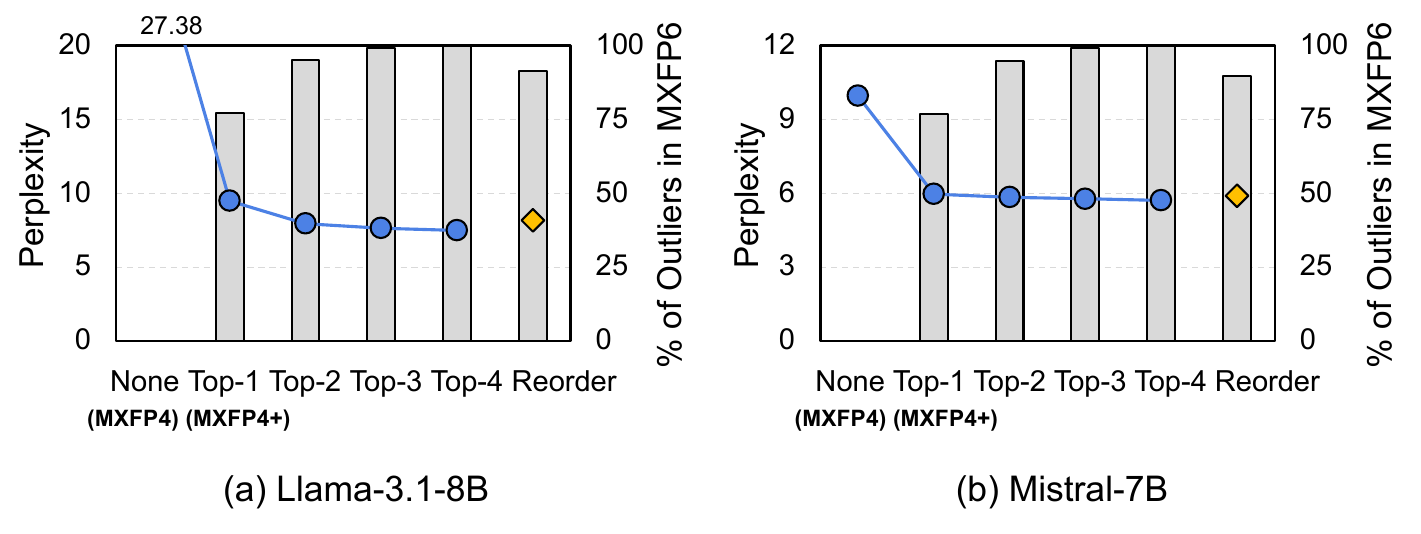}
  \vspace{-0.25in}
  \caption{
           Perplexity when representing \textit{top-k} magnitude elements of
           each block in MXFP6 while others in MXFP4 in a direct-cast setting.
           Bar plot represents the percentage of outliers in MXFP6.
          }
  \vspace{-0.15in}
  \label{fig:multi-bm}
\end{figure}

\putssec{reorder}{Addressing Multiple Outliers in a Block} 

Although we keep the proposed algorithm simple to minimize end-user overhead and
enable seamless integration with various frameworks, model performance can be
further improved when \name{} is optimized to capture outliers co-locating in
the same blocks. 

\myparagraph{Outlier Analysis.}
We represent the \textit{top-k} magnitude elements in MXFP6 for each block while
keeping others in MXFP4. \figref{multi-bm} shows perplexity and the
percentage of outliers represented in MXFP6 within activation tensors. 
We identify outliers using the 3$\sigma$ rule following~\cite{guo:tan23} and
focus on activations, as weights generally have a lower impact on model
performance.
The results show that extending MX+ to store additional BM index to track up to
two outliers provides some gain, while additional outlier representation shows
diminishing returns. 
This is because most activation outliers are represented in higher precision at
\textit{top-2}. 
To balance complexity and model performance, we choose channel reordering as an
optional optimization on top of MX+ to \textit{explicitly} separate outliers in
the same block.
When applying channel reordering with MX+, the perplexity and percentage closely
follow that of \textit{top-2} as shown in \figref{multi-bm}, which we discuss in
detail in the remaining section. 

\begin{table}[t]
\caption{
         Direct-cast inference accuracy on lm-eval-harness tasks.
         \textit{Reorder} denotes MXFP4+ with channel reordering applied to
         query and key matrices.
        }
  \vspace{-0.10in}
  \centering
  \resizebox{\columnwidth}{!}{%
    {\fontsize{14}{16}\selectfont
    \renewcommand{\arraystretch}{1.20}
      \begin{tabular}{llcccccc}
        \toprule[1.2pt]
        \textbf{Model}    		         &\textbf{Scheme}  &\multiline{\textbf{ARC}}{\textbf{easy}} 
                                                         &\multiline{\textbf{ARC}}{\textbf{challenge}} 
                                                         &\multiline{\textbf{Lam-}}{\textbf{bada}} 
                                                         &\multiline{\textbf{College}}{\textbf{CS}} 
                                                         &\multiline{\textbf{Int.}}{\textbf{law}} 
                                                         &\multiline{\textbf{Juris-}}{\textbf{prudence}}   \\
        \midrule
        \multirow{2}{*}{Llama-3.1-8B}  &MXFP4+           &70.29  &45.65  &66.83  &38.00  &66.12  &54.63    \\
        \cmidrule{2-8}                                                                                                  
                                       &\textit{Reorder} &72.77  &46.08  &68.58  &42.00  &77.69  &62.04    \\
        \midrule
        \multirow{2}{*}{Mistral-7B}    &MXFP4+           &74.20  &47.78  &70.79  &45.00  &67.77  &65.74    \\
        \cmidrule{2-8}                                                                                         
                                       &\textit{Reorder} &75.42  &48.55  &71.61  &49.00  &69.42  &66.67    \\
        \bottomrule[1.2pt]
      \end{tabular}
    }
  }
  \label{tab:outlier_scatter}
  \vspace{-0.10in}
\end{table}

\myparagraph{Scattering Outliers with Channel Reordering.}
As shown in \figref{act-analysis-2}, activation outliers are typically
concentrated at the channel granularity.
To allow more outliers to be identified as BM, we can also \textit{explicitly}
scatter outliers across blocks via channel-wise reordering.
For instance, we first sort the channels of each activation based on the number
of outliers.
Channels with the most outliers are then placed one in every 32 (i.e., block
size) channels. The remaining sorted channels are split in half, and we arrange
the lower half channels in the remaining places in descending order, followed
by the upper half channels in the same manner.

\tabref{outlier_scatter} shows the accuracy results for channel-wise
reordering, denoted as \textit{Reorder}.
The improvement stems from more precise outlier representations.
For example, the percentage of blocks with multiple outliers among
outlier-containing blocks decreases from 22.52\% to {4.58\%} in a sampled query
matrix after reordering.
For each task, we predetermine the channel ordering of query and key matrices
by averaging outlier counts per channel between the two matrices using 10\% of
samples.
Both matrices use identical channel ordering to maintain mathematical
correctness.
Reordering is fused with quantization by storing each quantized output at its
reordered channel address, making the reordering overhead negligible.

\putsec{related-work}{Related Work}

\myparagraph{Block-Based Data Formats for DNNs.}
Block floating point (BFP) has been adopted for efficient inference and
training of DNNs in academia~\cite{
zha:mcd22,lo:liu23,kim:oh24,det:pag23,lo:lee23,dru:lin18,yeh:ste22,dai:ven21,zha:che23}.
FAST~\cite{zha:mcd22} uses different BFP precisions for different layers and
training iterations during DNN training.
Bucket Getter~\cite{lo:liu23} implements intermediate accumulators in BFP PEs,
each of which accumulates values within a similar exponent range.
In recent years, industry has also increasingly explored BFP variants for
efficient DNN processing~\cite{fow:ovt18,dar:lo20,smx,kos:web17}. 
However, defining different data formats across organizations increases
end-user overhead and software fragmentation.
To address this, multiple companies collaborated to standardize data formats
and introduced Microscaling formats~\cite{mx-ocp}, which are increasingly
integrated into computing systems via software and hardware
support~\cite{blackwell,amd-mi355x,intel-xeon6}.
\name{} builds on this standard with a non-intrusive extension, making it
easily deployable across a wide range of platforms.

\begin{table}[t]
  \caption{Comparison of different quantization schemes.}
  \vspace{-0.10in}
  \resizebox{\linewidth}{!} {
    {\fontsize{10}{12}\selectfont
    \renewcommand{\arraystretch}{1.00}
      \begin{tabular}{l|ccccccc|c}
        \toprule[1pt]
        \textbf{{Scheme}}                              & AWQ   & SqueezeLLM & SMQ & QuaRot 
                                                       & OliVe & Tender & LLM-FP4 & \textbf{\name{}} \\
        \midrule
        \textbf{\multilineleft{Compute}{Efficiency?}}  &\low  &\low  &\high &\high 
                                                       &\high &\high &\high &\textbf{\high} \\
        \midrule
        \textbf{\multilineleft{Standard}{\& General?}} &\high &\high &\high &\high 
                                                       &\low  &\high &\low  &\textbf{\high} \\
        \midrule
        \textbf{\multilineleft{High}{Accuracy?}}       &\high &\high &\low  &\low 
                                                       &\low  &\low  &\low  &\textbf{\high} \\
        \bottomrule[1pt]
      \end{tabular}
    }
  }
  \label{tab:quantization_comparison}
  \vspace{-0.20in}
\end{table}

\myparagraph{Outlier-Aware Quantization.}
Efficient execution of DNN workloads has been extensively studied in various
contexts~\cite{che:eme16,jou:kur23,rea:wha16,qin:sam20,din:lia17,ven:swa21,
lee:lee24,kwo:li23,zha:zho23}.
Quantization is one of the most widely used strategies for achieving efficient
DNN execution~\cite{zha:liu21,yu:pra24,zad:mah22,han:xin16,tam:hoo21}, and
several previous studies focus on preserving the precision of outliers, which
is critical for maintaining model accuracy.
In the context of accelerating convolutional neural networks,
OLAccel~\cite{par:kim18} implements both 16-bit and 4-bit MAC units, with the
16-bit units handling outlier computations. DRQ~\cite{son:fu20} employs an
algorithm to identify accuracy-sensitive regions in tensors and performs
high-precision computation on these regions. 
While effective, these approaches rely on mixed-precision computation, which
leads to increased hardware complexity and unaligned memory access.

\tabref{quantization_comparison} compares \name{} with existing outlier-aware
work for LLMs. 
AWQ~\cite{lin:tan23} and SqueezeLLM~\cite{kim:hoo24} are weight quantization
methods that perform computation in high precision after dequantization.
SmoothQuant (SMQ)~\cite{xia:lin22} and QuaRot~\cite{ash:moh24} reduce the
magnitude of activation outliers through rescaling or rotation but show subpar
accuracy at low bit widths because outliers are not completely addressed.
OliVe~\cite{guo:tan23} employs custom data formats, whereas \name{} is based on
standard formats while providing higher accuracy.
Tender~\cite{tender} groups activation channels with similar dynamic ranges and
achieves comparable accuracy to MXFP4, as discussed in the paper.
LLM-FP4~\cite{liu:liu23} uses a custom floating-point format with channel-wise
scales encoded into the exponent bias. We observe that LLM-FP4 performs worse
than MXFP4 in our experiments. 

\putsec{conclusion}{Conclusion}

Serving LLMs requires substantial compute and memory resources, and the MX data
formats developed by leading industry companies are increasingly being adopted
to mitigate these challenges.
In this work, we investigate the implications of employing MX formats for LLM
inference and identify that model performance significantly degrades under
ultra-low precision due to quantization of activation tensors containing
outliers.
Building on the insight that the block absolute maximum element does not need
to store its exponent in the element data type, we propose \name{}, a
non-intrusive extension to MX that repurposes the exponent field as an extended
mantissa.
Without adding complexity, \name{} substantially improves model performance
across various precisions, with greater gains at lower bit widths.
It also enables straightforward deployment in software integration scenarios
with marginal overhead during inference, which can be virtually eliminated with
hardware support.

\begin{acks}

  This work was supported by the Institute of Information \& Communications
  Technology Planning \& Evaluation (IITP) grants funded by the Korean
  government (MSIT) (IITP-2025-RS-2022-00156295, IITP-2025-RS-2023-00256081)
  and a research grant from Samsung Advanced Institute of Technology (SAIT).
  The Institute of Engineering Research at Seoul National University provided
  research facilities for this work. Jaewoong Sim is the corresponding author.

\end{acks}


\bibliographystyle{ACM-Reference-Format}
\balance
\bibliography{refs}

\end{document}